\documentclass[final, journal]{IEEEtran}
\usepackage{amsmath,amsfonts}
\usepackage{algorithmic}
\usepackage{array}
\usepackage[caption=false,font=normalsize,labelfont=sf,textfont=sf]{subfig}
\usepackage{textcomp}
\usepackage{stfloats}
\usepackage{url}
\usepackage{multirow}
\usepackage{csquotes}
\usepackage{bm}
\usepackage{threeparttable}
\usepackage{makecell}
\usepackage[switch]{lineno} 
\usepackage[most]{tcolorbox}
\usepackage{amssymb}
\usepackage{verbatim}
\usepackage{xcolor}
\usepackage{tikz}
\usetikzlibrary{calc}
\usepackage{graphicx}
\graphicspath{{fig/}}
\DeclareGraphicsExtensions{.pdf,.jpeg,.png}
\usepackage{cite}
\def\BibTeX{{\rm B\kern-.05em{\sc i\kern-.025em b}\kern-.08em
    T\kern-.1667em\lower.7ex\hbox{E}\kern-.125emX}}
\usepackage{balance}

\begin{document}
\title{Large-scale and Efficient Texture Mapping Algorithm via {Loopy} Belief Propagation}
\author{Xiao Ling, Rongjun Qin,~\IEEEmembership{Senior Member,~IEEE}
  \thanks{Xiao Ling is with Geospatial Data Analytics Laboratory, The Ohio State
    University, 218B Bolz Hall, 2036 Neil Avenue, Columbus, OH 43210, USA, and
    also with Department of Civil, Environmental and Geodetic Engineering, The
    Ohio State University, 218B Bolz Hall, 2036 Neil Avenue, Columbus, OH 43210,
    USA, also with College of Astronautics, Nanjing University Of Aeronautics
    And Astronautics, 29 Jiangjun Avenue, Jiangning District, Nanjing, Jiangsu 211106,
    China. (email: xlingsky@nuaa.edu.cn)}
  \thanks{Rongjun Qin is with Geospatial Data Analytics Laboratory, The Ohio State
    University, 218B Bolz Hall, 2036 Neil Avenue, Columbus, OH 43210, USA, and
    also with Department of Civil, Environmental and Geodetic Engineering, The
    Ohio State University, 218B Bolz Hall, 2036 Neil Avenue, Columbus, OH 43210,
    USA, also with Department of Electrical and Computer Engineering, The Ohio
    State University, 205 Dreese Lab, 2036 Neil Avenue, Columbus, OH 43210, USA,
    also with Translational Data Analytics Institute, The Ohio State University.
    (email: qin.324@osu.edu, corresponding author)}}

\markboth{Preprint version \newline To appear in Journal of Transactions on Geoscience and Remote Sensing(2023)}%
{Large-scale and Efficient Texture Mapping Algorithm via Parallelized Loopy Belief Propagation}

\maketitle

\begin{abstract}
  Texture mapping as a fundamental task in 3D  modeling has been well established for well-acquired aerial assets under consistent illumination, yet it remains a challenge when it is scaled to large datasets with images under varying views and illuminations. A well-performed texture mapping algorithm must be able to efficiently select views, fuse and map textures from these views to mesh models, at the same time, {achieve} consistent radiometry over the entire model. Existing approaches achieve efficiency either by limiting the number of images to one view per face, or simplifying global inferences to only achieve local color consistency. In this paper, we break this tie by proposing a novel and efficient texture mapping framework that allows the use of multiple views of texture per face, at the same time to achieve global color consistency. The proposed method leverages a {loopy} belief propagation algorithm to perform an efficient and global-level probabilistic inferences to rank candidate views per face, which enables face-level multi-view texture fusion and blending. The texture fusion {algorithm}, being non-parametric, brings another advantage over typical parametric post color correction methods, due to its improved robustness to non-linear illumination differences. The experiments on three different types of datasets (i.e. satellite dataset, unmanned-aerial vehicle dataset and close-range dataset) show that the proposed method {has} produced visually pleasant and texturally consistent results in all scenarios, with an added advantage of consuming less running time as compared to the state of the art methods, especially for large-scale dataset such as satellite-derived models.
\end{abstract}

\begin{IEEEkeywords}
  texture mapping, belief propagation, multiple labels, image blending.
\end{IEEEkeywords}

\section{Introduction}
\IEEEPARstart{T}{exture} mapping (TM) is defined as assigning textural materials to 3D geometry
(meshes or polyhedral models), such that it is visually consistent and
contextually correct. It serves as a fundamental step for almost all 3D modeling
pipelines, and {this is especially} becoming a standard step in 3D mesh modeling
using well-acquired aerial assets \cite{gruen_operable_2020,
  hoegner_automatic_2016, guidi_multi-resolution_2009}. However, this is still a
challenging problem when such a task is scaled to a very large {dataset}, or in
general scenarios where the image size is large and {images are} captured under
drastically different lighting conditions. For example, TM using multi-view
satellite images \cite{qin_rpc_2016, qin_automated_2017} possesses {both challenges.} On the one hand, performing TM on mesh models at the city scale
requires the TM algorithms to balance both memory and computational efficiency;
on the other hand, these multi-images are often collected at different
times/seasons under various illumination. Such challenges similarly exist when
sources of textures are coming from different sensors. 

There exists a large number of well-performed TM algorithms in both the computer
graphics and {computer vision domains} \cite{waechter_let_2014,lempitsky_seamless_2007,bi_patch-based_2017}, aiming at image
selection, and color balancing among neighboring {triangles/faces} locally, and
color/lighting consistency among the entire 3D model. {Oftentimes, the process of
performing global color balancing requires highly efficient algorithms.
These algorithms, however, come at the expense of sacrificing the used information among
all eligible texture materials from different viewpoints. For example, the
classic graph-cut based image selection algorithm used in \cite{waechter_let_2014,lempitsky_seamless_2007}, considered selecting a single view
for each face/triangle, while attributed solutions of addressing illumination
disparities among different views solely to post-correction algorithms, which by
concept, is limited by the capacity of the correction models; for instance,
major post-correction models utilize either a bias term or at maximum a linear correction model, which are incapable of
addressing non-linear color differences that are otherwise very common in
practice.} On the other hand, the very classic, view-dependent TM scheme
\cite{callieri_masked_2008}, considers a fusion of multiple image textures seen by a
single face/triangle in the 3D object, using weighting strategies based on the
viewing angle, the level of coverage, etc. However, since the fusion is
carried out independently for each triangle/face in the 3D model, {this,} in
contrast, sacrifices the global color consistencies among neighboring faces and
the overall 3D model, {for two reasons}: 1) the fusion for individual
{triangles/faces} will consume a large number of computation power (considering
models with millions of triangle faces and more); 2) the fused image lost its
belongingness to a particular source view, thus is unable to exploit the level
of color consistencies among neighboring triangle/faces as if they were from a
single image. If these two seemingly conflicting schemes can be leveraged {under the same TM framework,}
it will facilitate TM algorithms that enjoy the pros
of existing TM methods, to be thus both accurate and efficient.  

In this paper, we propose a novel texture mapping framework to achieve both
efficiency and accuracy as to address the aforementioned limitations for TM:
instead of fusing the multiple image textures at the first place, we consider
ranking these multiple image textures at the triangle level, which {leverages} a
number of cues {to yield} consistent textures and only fuse the top few. The ranking
of image texture for each triangle is {obtained} by utilizing not only local
cues, such as the viewing angles, {and} coverage of the image, but also global
cues such as the color resemblance of image texture candidates and consistency
in their source image ID for neighboring {triangles/faces}. This framework will
additionally allow optional post-correction as needed, or directly involve these
multiple image textures in the post-correction process. In other words, our
method, instead of selecting a single view, {preserves} multiple image textures prior
to fusion and post-correction, which will, on one hand, allow global inferences
to select multiple image texture candidates, and on the other hand, {provide} the
flexibility to perform non-parametric color corrections to accommodate various
{levels} of illumination differences among different images. We found the solution
of our proposed method can be achieved through a belief-propagation framework
\cite{yanover_finding_2003}, owing to its unique advantage in allowing global
probabilistic inferences through information passing for multiple variables
(here referred to as multiple image texture candidates), as compared to graph-cut
solutions \cite{kolmogorov_what_2004,paragios_graph_2006} that only offer a
deterministic solution as the cut. We carry out experiments using various
datasets including satellite, aerial and close-range images and models using our
method and compared it with {the state-of-the-arts (SOTAs)} and have shown unique
{advantages in} various case scenarios. As for example, because our approach allows
{reserving} multiple texture candidates, it is able to generate consistent and
visually pleasing textures for fa\c{c}ades in high-altitude mapping cases (i.e.,
satellite images) despite that fa\c{c}ade information is rather limited. 

The rest of this paper is organized as follows: \textbf{Section \ref{sec:related_work}} presents a literature
survey related to existing methods in TM for 3D modeling. \textbf{Section \ref{sec:methodology}} describes
our proposed algorithms and {especially our critical additions} to the
existing literature. Experimental results and comparative studies on three
different datasets are presented in \textbf{Section \ref{sec:experiment}}. \textbf{Section \ref{sec:conclusion}} concludes this paper by
analyzing the advantages and drawbacks of our methodology that inform our
planned future works. 

\section{Related work}\label{sec:related_work}
\noindent Given a set of images with poses and a 3D geometric model (e.g. meshes
or polyhedral models), a plain version of a TM algorithm transverses all the
faces, and for each face, it selects a visible view and {projects} its valid
portion to the face as an assigned texture material. This plain version of TM
algorithm apparently drives two practical challenges in order to achieve
optimized results: first, how to select the view(s) among multiple (often a
large number of) visible views to maximize texture quality; second, texture
materials from different source views on neighboring faces inevitably result
{in inconsistent} texture transitions, and how to correct such texture inconsistency
across the entire model. These two challenges are correlated, as consistent
texture transitions may serve as a criterion for view selection, e.g., one can
purposely select views to minimize inconsistent source views for textures of
neighboring views. Therefore, the TM {has became} a multi-complex problem that
requires solutions to be systematic, robust and efficient.   

Existing approaches either focus on addressing one of the challenges, or both
following a sequential order. Generally, we categorize them into two classes: 1)
the blending-based methods and 2) {the Markov Random Field} (MRF) based TM framework. The former mainly focuses on addressing the second challenge by exploiting weight fusion of all possible image textures \cite{callieri_masked_2008}, to achieve texture homogeneity across the entire model, and the latter, however, starts with view selection by modeling views per face as a random variable in a Markov Random Field (MRF) \cite{li_markov_1994}, thus to allow the consistency constraints to be imposed on neighboring faces to achieve a global texture consistency optimum \cite{lempitsky_seamless_2007,xu_an_2011}. Such an approach is often followed by an optional post-correction of the remaining inconsistent textures \cite{waechter_let_2014,lempitsky_seamless_2007}. 

\subsection{Blending-based Methods}
\noindent The blending-based approaches
\cite{callieri_masked_2008,grammatikopoulos_automatic_2007,hoegner_automatic_2016}
consider all visible views per face, and project all input images onto the
surface of the 3D model using the camera pose and then blend/fuse them. The
blending is achieved either with equal weights or with other image quality
indicators, such as resolution, viewing angles etc., to achieve a consistent
texture map. The underlying concept is that such a weighted sum inherently
builds a spatially smooth function with respect to the image textures, and as a
result the blended image {averages out} potential conflicting illuminations
of different images, to achieve smooth transitions across neighboring
{faces}. Such blending methods, called alpha blending with many of its
variants \cite{allene_seamless_2008,baumberg_blending_2002,li_fast_2006}.
Specifically, a commonly used approach, called distance transform based alpha
blending \cite{szeliski_image_2006}, is to firstly compute the distance to the
nearest invalid pixel for each pixel on {an} image to generate {a}
distance map and then fuse all views together with distance maps as weights,
which inherently assume {the} image texture {resides} in the center of the image with higher weights and vice versa.

This type of approach, however, has some critical drawbacks. First, the fact of
fusing all visible views means {tolerating} all potential {sources} of
errors which are not easily accountable by weighted blending, such as drastic
image resolution difference, camera pose errors, strong illumination differences
etc., thus often result in blurred texture materials, reduced textured
resolution and ghosting artifacts. second, although this type of approach enjoys
the benefit of memory efficiency due to that it is a local method (each face is
processed independently), blending all visible views also means to tolerate
notoriously heavy computation when the number of views is large (in the order of
tens to hundreds), leading to sometimes unbearable waiting time. Third, this
type of approach being local, although {produces} locally smooth texture
transitions, will not accommodate global illumination differences that are
spatially distant. It should be noted there are approaches to deal with the
first drawback by introducing more advanced approaches than simple weighted
blending{. For example}, Bi et al, \cite{bi_patch-based_2017} improved the
blending method by synthesizing aligned images from input images even with
inaccurate camera poses by proposing a patch-based optimization system to avoid
ghosting and blurring artifacts in the result. {Despite} the performance
gain, this approach with an added layer of complexity will only further increase
the computation when dealing with a large number of views{. Thus, it} is only suitable for close-range and small-range models, such as indoor scenes and small objects.

\subsection{MRF-based Texture Mapping framework}
\noindent Lempitsky and Ivanov \cite{lempitsky_seamless_2007} first proposed the
MRF based TM framework to select an optimal view for each mesh face. Based on
their work, other similar works
\cite{allene_seamless_2008,gal_seamless_2010,waechter_let_2014,fu_texture_2018,li_fast_2018}
focused on improving the terms of the energy function or {modifying} them to
{adapt to different scene contexts,} such as for RGB-D (depth) data . Among
these methods, \cite{waechter_let_2014} plays an important role for large-scale
TM by introducing high performance data term and smooth term as well as a
post-correction on texture colors with a two-step adjustment (i.e. global seam
leveling, followed by a local adjustment with Poisson editing
\cite{perez_poisson_2003}) to achieve high efficiency and uniform color. It is
also one of the most well-engineered and open-source {methods} adopted by the
well-known OpenMVS package \cite{cernea_openmvs_2015}. Therefore, it
consistently serves as a baseline approach for comparison, and {is still} presented as one of the {top approaches} for dealing with practical TM problems with large datasets. However, this type of approach still suffers from a few drawbacks: 1) The fact of only selecting a single view per face will inevitably result in sub-optimal texture selection at certain faces{. For example,} the {faces of building roofs and fa\c{c}ades} are topologically connected and ideally {chosen} the same view for both will avoid seamlines between two faces, while the same view is apparently not ideal to provide the best textural coverage, as roof faces favor nadir images while fa\c{c}ade faces favor oblique views. Thus, in theory, it is very difficult to design a view selection algorithm that {accommodates} multiple objectives. 2) the fact of using only one view will inevitably encounter scenarios where one needs to consider post-correction of inconsistent texture colors at faces which are beyond what a single view can cover. Additionally, the post-correction of inconsistent textures is essentially a global operation, which requires heavy computation even with simple linear (or bias) corrections. In Waechter et al. \cite{waechter_let_2014}, this was performed two-step method to correct complex color inconsistencies, while it takes a long {running time}, is still insufficiently accurate to yield realistic colors, e.g. issues of discoloration affected by shadowed fa\c{c}ades connected with well illuminated roof tops. 

\textbf{MRF Solver:} The MRF framework embedding various constraints, such as
number of views, coverages and incidence angles etc., can be solved by various
classes of solvers such as {Loopy Belief Propagation (LBP), Graph Cuts (GC)} and Convergent Tree Reweighted Message Passing (TRW-S) \cite{kolmogorov_convergent_2005}. According to Lempitsky and Ivanov \cite{lempitsky_seamless_2007}, GC solver demonstrates the best performance, which was further used by other similar works \cite{allene_seamless_2008,waechter_let_2014,fu_texture_2018,li_fast_2018}. As a result, the fact of these methods only returning a single view per face, is limited by the GC solver, because it only provides a final labeling (which view to choose) subject to minimizing the formulated energy function. 

\textbf{Post-correction of inconsistent textures:} An example of the
post-correction method \cite{lempitsky_seamless_2007}, computed globally optimal
luminance correction as a bias term added to each vertex of faces, with the goal
to minimize the luminance differences per vertex at seams to allow smooth color
transition. Waechter et al. \cite{waechter_let_2014} further improved it by
adjusting the color along the seams via Poisson editing to allow smoother
transitions {across} neighboring faces. The solver of these bias terms will
result {in} a large and exponentially growing sparse matrix with the number of faces, facing both computation and numerical stability challenges.

However, GC cannot be parallelized easily due to the graph’s irregular
structure, while the LBP solver can be fully parallelized, which implies
{the} LBP solver may overtake {the} GC solver on multi-core machines for performance.
More {importantly}, the LBP solver is able to return {the} costs of views for each mesh face,
so that not just one view per face but top N views can be assigned to a face at once.

\subsection{Considerations leading to the proposed method}\label{sec:related_work_proposed_method}
\noindent While the MRF framework is regarded as a popular approach, one
fundamental bottleneck is that it is limited to {selecting} one view per
face{, which leads} to sub-optimal views that cannot accommodate multiple objectives. If more
views are selected under the MRF framework, it may allow {flexibility} to enable
blending methods on these selected views and thus to achieve seamless texture
transitions without complex post-correction methods. We consider the major
hurdle of this \enquote{one view} limitation is caused by the use of
\enquote{GC} solver, although it is regarded as the most accurate (reported by
\cite{lempitsky_seamless_2007}){. We} therefore leverage this by using the LBP
solver to perform the MRF optimization, since it
allows explicit probabilistic inferences for the random variable, which we can
use a cue to design more flexible constraints to achieve multiple objectives, as
well as return multiple candidates of views based on the explicit probability
distributions for each potential view candidates. Therefore, the idea here is to
construct an energy function using constraints such as view consistency,
coverage, viewing angles, etc., to maximize the quality of the image candidate;
retrieve the {probability} distribution of potential views via a LBP solver, rank
them to reflect the multiple objectives, and only blend the top N views that
fulfill the objectives to achieve smooth texture transition (without including
bad texture that pullulates the blended image). This will automatically achieve
global color consistency (due to the MRF frame), and local smooth texture
transition (due to per face blending), without the possible need to perform
post-correction, which was supposed to alleviate color inconsistencies due to the
lack of blending in MRF frameworks.

\section{Our proposed approach}\label{sec:methodology}
\begin{figure*}[t]
  \centering
  \includegraphics[width=0.95\textwidth]{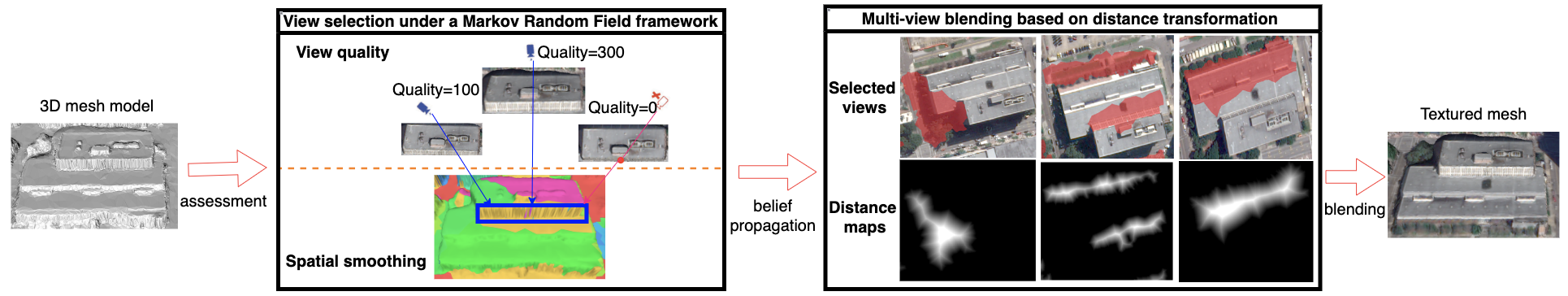}
  \caption{{The proposed workflow. Given a 3D mesh model and the
      corresponding oriented images, the qualities of views are firstly computed
      as visibility, projected area and {color-consistency} for each face,
      and then all these terms are formulated into a MRF energy function solved
      by LBP, which returns top N views to each face. A multi-view blending algorithm is finally applied to yield the seamless high-quality texture.}}\label{fig:workflow}
\end{figure*}

\noindent Following our considerations as mentioned in \textbf{Section
  \ref{sec:related_work_proposed_method}}, our proposed work follows the Markov
Random Field (MRF) based TM {framework,} which constructs a global
optimization involving data terms and smooth terms to balance the quality per
view and the severity of seams between texture patches. As shown in \textbf{Fig.
  \ref{fig:workflow}}, given the 3D mesh and its {associated} images with known
orientation parameters, the qualities of views are computed in terms of
visibility, projected area and {color-consistency} for each face, and then
all these terms are packed into a pairwise MRF energy formulation solved by
{Loopy Belief Propagation (LBP)} to assign top N best views to each face, finally a multi-view blending algorithm based on distance map \cite{fabbri_2d_2008} is applied to yield the texture with uniform color. Details of those steps are introduced in the following sub-sections. 
 
\subsection{{Multi-objectives for view quality assessment}}\label{sec:view_quality}
\begin{figure}[ht]
  \centering
  \includegraphics[width=0.8\columnwidth]{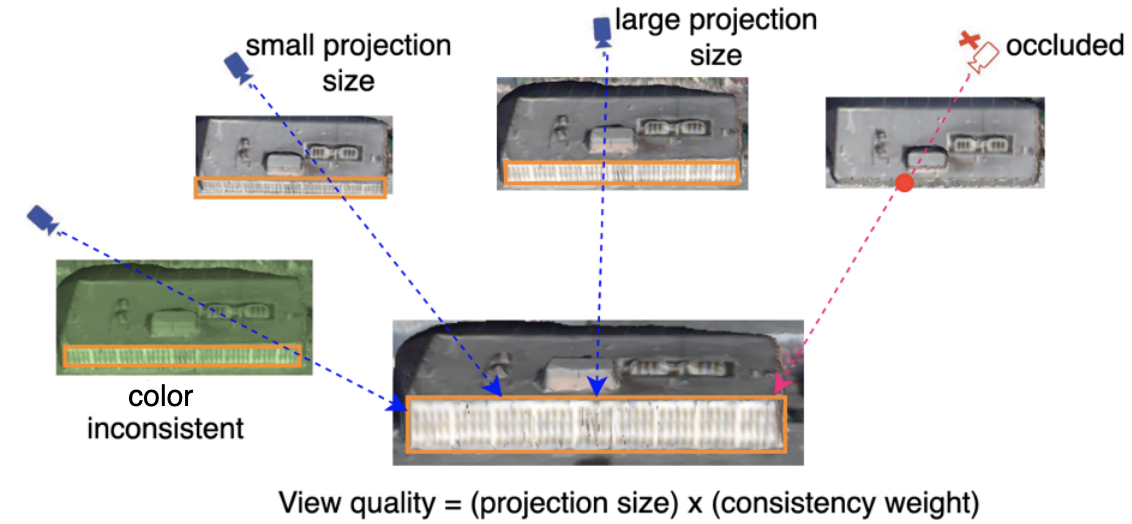}
  \caption{{Illustration for view quality assessment. All occluded views are
      eliminated firstly. The resolution of the projected area (i.e. the projection
      size) is used as a quality indicator since a larger projection size implies
      small incidence angle leading to less distorted texture and higher image
      resolution leading to more textural details. Color inconsistency is
      another quality indicator to warrant consistent texture illumination among
    neighboring faces.} }\label{fig:viewquality}
\end{figure}

\noindent To evaluate the quality of views when texturing a face,
we first select the visible views (by an occlusion detector), and then employ
quality indicators with the following considerations:
 1) the selected views should contain maximal texture details, 2)
 the selected view textures should have minimal geometrical
 distortions, and 3) the selected views should be \enquote{popular} to warrant
 consistent colors among neighboring faces. The first two considerations are
 coherent, and can be well-represented by the resolution of the projected areas \cite{allene_seamless_2008,buehler_unstructured_2001,zhou_selection_2021},
 meaning the number of pixels in the projected footprint of a triangle face in a
 view, because this inherently implies a smaller incidence angle leading to less
 distorted texture, and more pixels per unit area leading to more textural details.
\enquote{Popular} views refer to the fact of view selecting being smooth,
meaning that neighboring faces select views in the color-consistent set (i.e.
the majority), consistent texture
illumination can be achieved as they came from the majority. In summary, the
following quality indicators are considered when evaluating a view (visual
illustration can be found in \textbf{Fig. \ref{fig:viewquality}})


\textbf{(1) Visibility.} {Occluded views (non-visible views) cannot be considered as view candidates.} Here, to accommodate generic camera models (i.e. frame camera model, linear-array or parametric models (RPC (Rational Polynomial Coefficient)-based)), we use a simple z-buffer algorithm \cite{greene_hierarchical_1993} that renders depth values from the geometry to potentially candidate views and only views identified as visible will be considered. Such a rendering scheme can be hardware accelerated and present an efficient means to filter out invisible views.  

\textbf{(2) Resolution of the projected area} $\mathbf{S}$. As described above,
{a} larger projection area implies better {incidence} angle and higher
image resolution{. Thus,} the projection area of a face on a view would be an
efficient way to judge view quality. Although, as argued in Waechter’s paper
\cite{waechter_let_2014}, optimizing the projection size may lead to
out-of-focus blur as it tends to select views whose image plane are close to the
object. This is, however, only problematic {in} extremely close-range
{applications}, which can be easily leveraged {by} other constraints (e.g. the distance between the perspective center to the object's surface).

\textbf{(3) {Color consistency}} $\mathbf{\omega}$. To minimize the inconsistency
of final texture charts, views with consistent colors should be given higher
{weights} of consideration. {On the contrary, views with inconsistent colors
  against the majority (e.g. some shadowed textures) should be punished.}
Here, among all view candidates, we tend to first compute the mean brightness (color) of these views and give higher weight to those close to the mean and lower to drastically different ones. The mean color is computed via a modified mean-shift algorithm \cite{waechter_let_2014}, computed as: given a list of views for a specific face,
\begin{enumerate}
  \item Compute the face projection’s mean color $c_i$ for all visible pixels on view $l_i$.
  \item Compute mean $\mu$ and covariance matrix $\sum$ of all views’ mean color $c_i$.
  \item Evaluate the multi-variate Gaussian function $e^{\frac{-(c_i-\mu)^T \sum(c_i-\mu)}{2}}$ for each view $l_i$.
  \item Eliminate the views whose function value is below a threshold (we use 0.006 for 8-bit image).
  \item Repeat 3.-4. for 10 iterations or until the inversion of $\sum$ becomes stable, i.e. all entries of $\sum$ drop below $10^{-5}$, or the number of inliers is less than 4.
  \item Compute the Gaussian function value for each view $l_i$ as its weight of view quality to raise up the probability for {color-consistent} views in the following view selection procedure.
  \end{enumerate}

  In summary, the final score of view quality $Q$ of a view $l_i$ for a
  specific face $F_i$ is computed by a product of face projection’s size
  $S(F_i,l_i )$ (i.e. the number of visible pixels within the projected region)
  and its {color consistency’s} weight $\omega(F_i,l_i)$ (see \textbf{Equation \ref{equ:view_quality}}), to prevent the view with less visible pixels or larger color difference being selected.

\begin{equation}
  \label{equ:view_quality}
  Q(F_i,l_i)=\omega(F_i,l_i )S(F_i,l_i)
\end{equation}

\subsection{View selection under a Markov Random Field framework}
\noindent The per-face view quality indicator as described in \textbf{Section
  \ref{sec:view_quality}} does not consider spatial smoothness,
{because a single/linear quality indicator for a face is insufficient to
  model the complex correlation between view quality and its potential
  smoothness in connection to its neighbors. It is possible that spatial
  smoothness contradicts the view quality, and on occasions, the selection will
  need to sacrifice the view quality (higher texture details), in order to
  achieve smooth color transitions}. Therefore, we consider optimizing
{the view selection} through a Markovian process using a MRF
Framework, which allows the consideration of {these quality indicators as well as}
view consistencies across spatially connected faces. Unlikely typical MRF solvers for view selection \cite{lempitsky_seamless_2007} which produce a single and finalized view per face, the aim of this proposed MRF process is to allow view indicator inferences to obtain top view candidates per face for texture fusion. The MRF energy function is formulated as follows:

\begin{equation}
  \label{equ:mrf}
  E(l)=\sum_{F_i\in \mathbb{N}_f}E_{data}(F_i,l_i)+\sum_{(F_i,F_j)\in \mathbb{N}_e}E_{smooth}(F_i,F_j,l_i,l_j)
\end{equation}
where, $l_i$ denotes the view assigned to a mesh face $F_i$, $\mathbb{N}_f$ is the mesh face set and $\mathbb{N}_e$ is the mesh edge set. The first term is the data term $E_{data}$ which prefers views with high quality for a face and the second term is the smoothness term $E_{smooth}$ which punishes inconsistent labels between neighboring faces which share a common edge. Details about these terms are further {given below}.

\begin{enumerate}
\item 	\textbf{Data term:} For the reasons described in \textbf{Section \ref{sec:view_quality}}, the cost $E_{data}(F_i,l_i)$ for a view $l_i$ assigned to a mesh face $F_i$ is computed as $E_{data}(F_i,l_i)=Q(F_i,l_i)$, here $Q(F_i,l_i)$ is the score of view quality computed from \textbf{Equation \ref{equ:view_quality}}. 
\item 	\textbf{Smoothness term:} The Potts model: $E_{smooth}=[l_i\ne l_j ]$
  ($[\cdot]$ is the Iverson bracket) is used as a smoothness term to favor
  {a} consistent label between neighboring mesh faces. This also prefers compact patches and is extremely fast to compute.
\end{enumerate}

\begin{figure}[ht]
  \centering
  \includegraphics[width=\columnwidth]{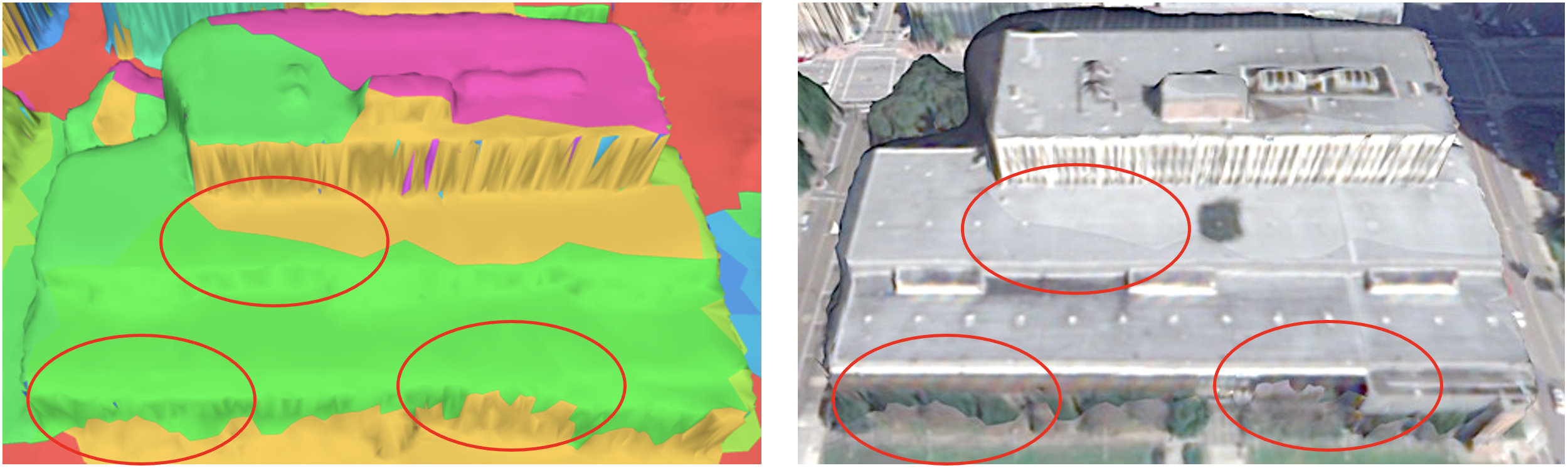}
  \caption{Luminance differences existing in singe (best) view based texture mapping. Left: One color represents a best view for each mesh face; Right: only the best view is assigned to map textures with no color adjustment. As highlighted in red circles, two observations are noted: 1) there are visible luminance differences between patches; 2) the fa\c{c}ade near the ground is textured by two views with apparently different qualities, one is favored by the roof and the other is favored by the ground.}\label{fig:luminance}
\end{figure}

The energy minimization problem (\textbf{Equation \ref{equ:mrf}}) can be solved
by many algorithms, e.g. {LBP, $\alpha$-expansion GC and Convergent Tree
  Reweighted Message Passing (TRW-S)}. However, for GC or TRW-S as typical
solvers, only return a single label, which will inevitably cause seamlines at
the edges shared by two faces with different labels, thus computationally
expensive post-correction is necessary, such as a local adjustment with Poisson
editing used in Waechter et al. \cite{waechter_let_2014}. Here we consider a
simple extension to avoid this computation: instead of assigning the best view
for each face, we reserve the top N (we used 3 in this work) views after the
inference and perform a simple and weighted linear blending. Therefore, we
apply a LBP solver to resolve the energy function (\textbf{Equation
  \ref{equ:mrf}}), and rank the candidates based on the cost volumes for each
possible {view}, which provides much more flexibility to blend textures to
potentially avoid seamlines. The diversity of views is depicted in \textbf{Fig.
  \ref{fig:luminance} (left)}, with the right image of \textbf{Fig.
  \ref{fig:luminance}} {showing} the seamlines if only a single (and the best view
is selected), and the blended result is shown in \textbf{Fig.
  \ref{fig:view_blending}(d)}. Since the data term gives lower costs to the
majority of views and the smoothness term prefers {a} consistent label between
neighboring faces, the color difference of the top N candidates for a face will
drop into a fine level and the view candidates for the neighboring faces will
have common elements as well after optimizing \textbf{Equation \ref{equ:mrf}}.
Details of the LBP solver {are} as follows:

Given a pair of neighboring nodes (i.e. two mesh faces $F_i$ and $F_j$ with a
common edge), the directed {message} $m_{F_i,F_j}$ from $F_i$ to $F_j$, which is
initialized to 1, is updated by considering all {messages} flowing into $F_i$
(except for message from $F_j$) via \textbf{Equation \ref{equ:message}}, where
$\mathbb{N}(F_i )\backslash F_j$ is the set of neighboring faces except $F_j$
for $F_i$, the cost of each view $l_i$ for face $F_i$ are computed from
\textbf{Equation \ref{equ:cost}} and the smallest cost is taken as its selected
view, this procedure iterates for a fixed number of iterations (e.g. 50 in our
experiments) or until \textbf{Equation \ref{equ:mrf}} reaches the minimum, and
views with the smallest cost for each {face} are taken as final solutions. In
practice, one usually normalizes the messages $m_{F_i,F_j}$ to sum to 1, so that
$\sum_{l_j}m_{F_i,F_j} (l_j)=1$, in every iteration for numerical stability.
Both messages (\textbf{Equation \ref{equ:message}}) and costs (\textbf{Equation
  \ref{equ:cost}}) computation are parallelizable {and have been fully parallelized in our implementation}.

\begin{equation}
  \label{equ:message}
  \begin{split}
    m_{F_i,F_j}^{new}(l_j )=\sum_{l_i}\Big(&E_{data}(F_i,l_i ) E_{smooth }(F_i,F_j,l_i,l_j )\\
    &\prod_{F_k\in \mathbb{N}(F_i )\backslash F_j}m_{F_k,F_i}^{old }(l_i )\Big)
  \end{split}
\end{equation}

\begin{equation}
  \label{equ:cost}
  C_{F_i } (l_i )=E_{data}(F_i,l_i ) \prod_{F_k\in \mathbb{N}(F_i )}m_{F_k,F_i } (l_i )
\end{equation}

The symbol N defines the maximum of qualified view candidates to be assigned to
a face, i.e., some faces may have less than N associated views after view
selection, since 1) there may be no sufficient views for a specific face due to
occlusion; 2) the final costs of some views are much larger than others’, which
may occur for the fa\c{c}ades viewed by a limited number of \enquote{good} views but many
\enquote{bad} views. For the second case, a ratio test is applied to eliminate the \enquote{bad}
views: assume $c=c_1,c_2,\cdots,c_N$ sorted in ascending order is the final
costs of top N candidates $l=l_1,l_2,\cdots,l_N$ for a specific face,
candidates $\{l_i,\cdots,l_N\}$ will be removed if $c_{i-1}\slash c_i$ is below a threshold (0.4 is used throughout our experiment as an empirical value).

\subsection{Multi-view blending based on distance transformation}
\begin{figure}[ht]
  \centering
  \footnotesize
  \begin{tabular}{@{}c@{}c@{}c}
                                                     \includegraphics[width=0.3\columnwidth,height=0.3\columnwidth]{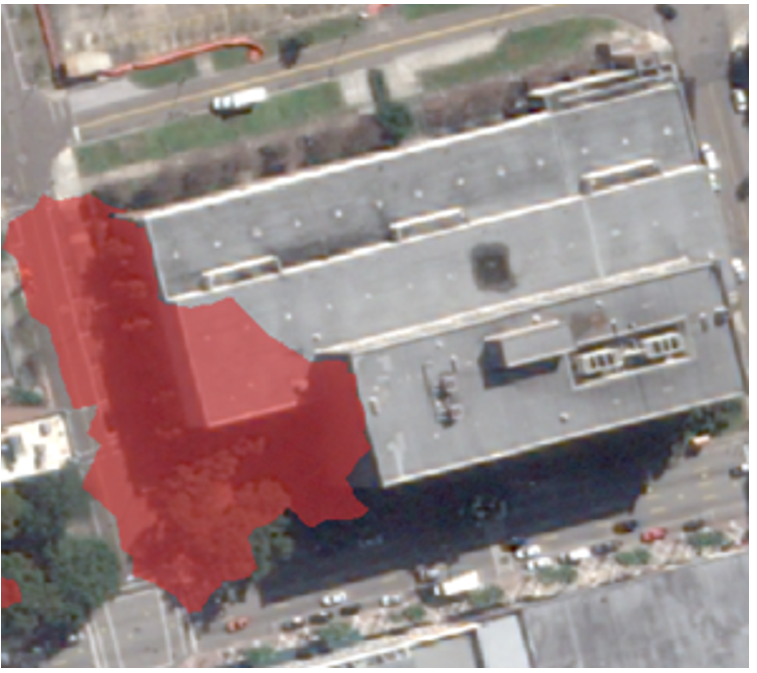}\;\;&  \includegraphics[width=0.3\columnwidth,height=0.3\columnwidth]{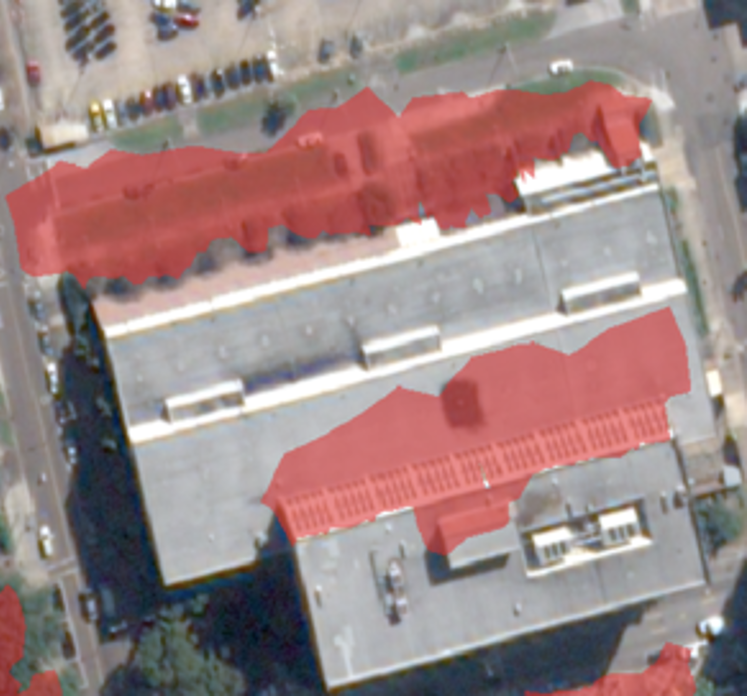}\;\;& \includegraphics[width=0.3\columnwidth,height=0.3\columnwidth]{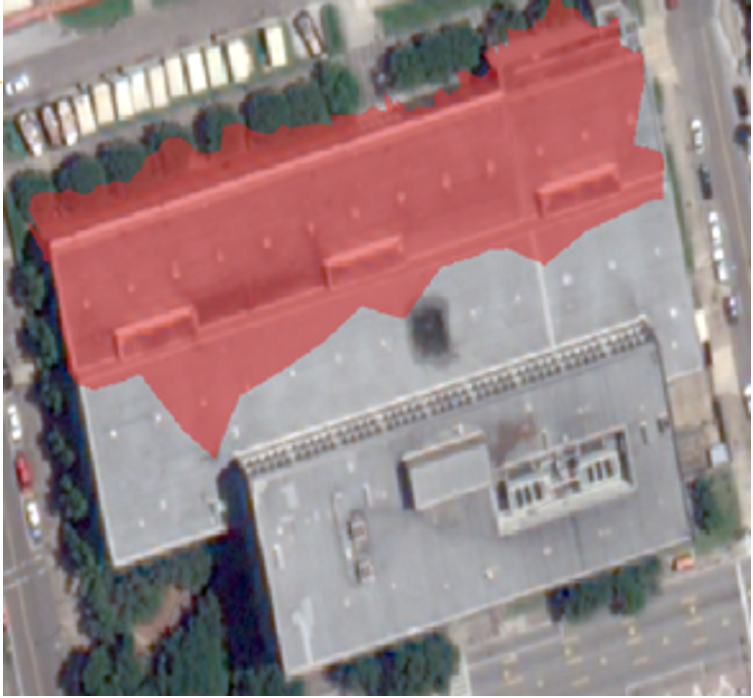}\\
                                                       \includegraphics[width=0.3\columnwidth,height=0.3\columnwidth]{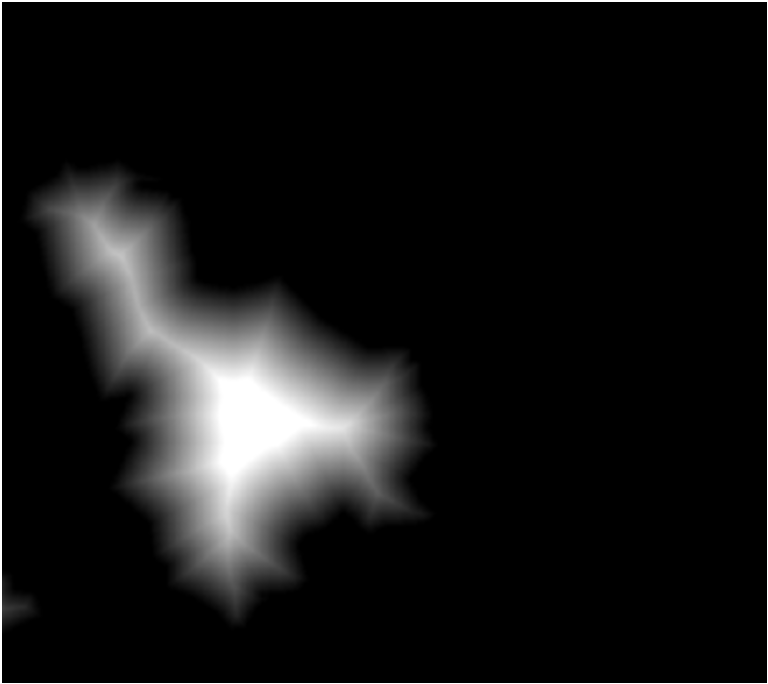}\;\;& \includegraphics[width=0.3\columnwidth,height=0.3\columnwidth]{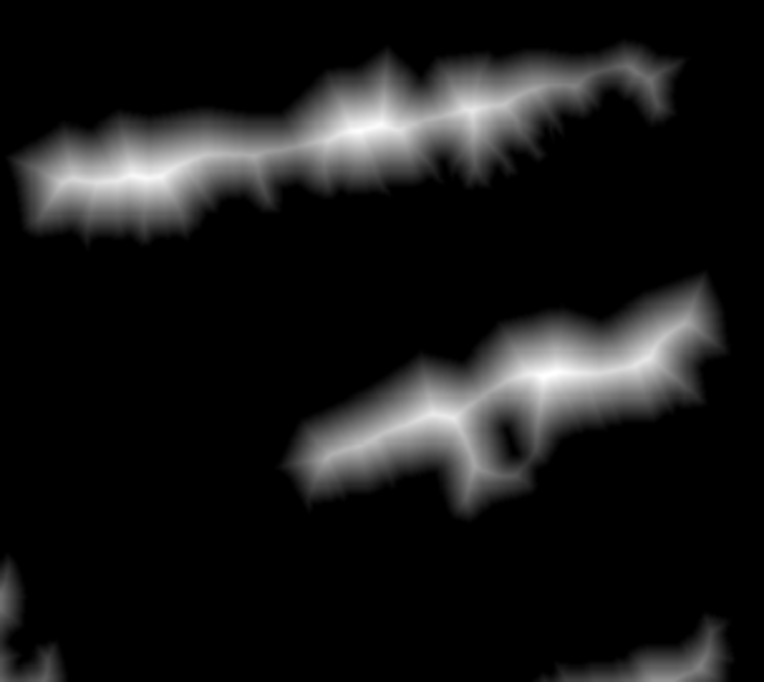}\;\;&   \includegraphics[width=0.3\columnwidth,height=0.3\columnwidth]{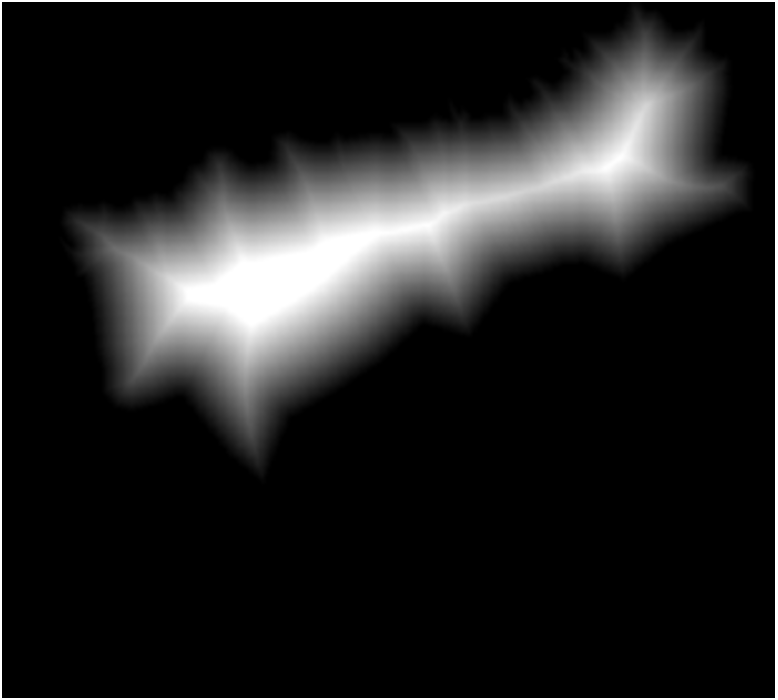}\\
    \multicolumn{3}{c}{(a)}
  \end{tabular}
  \begin{tabular}{@{}c@{}c}
    \includegraphics[width=0.49\columnwidth]{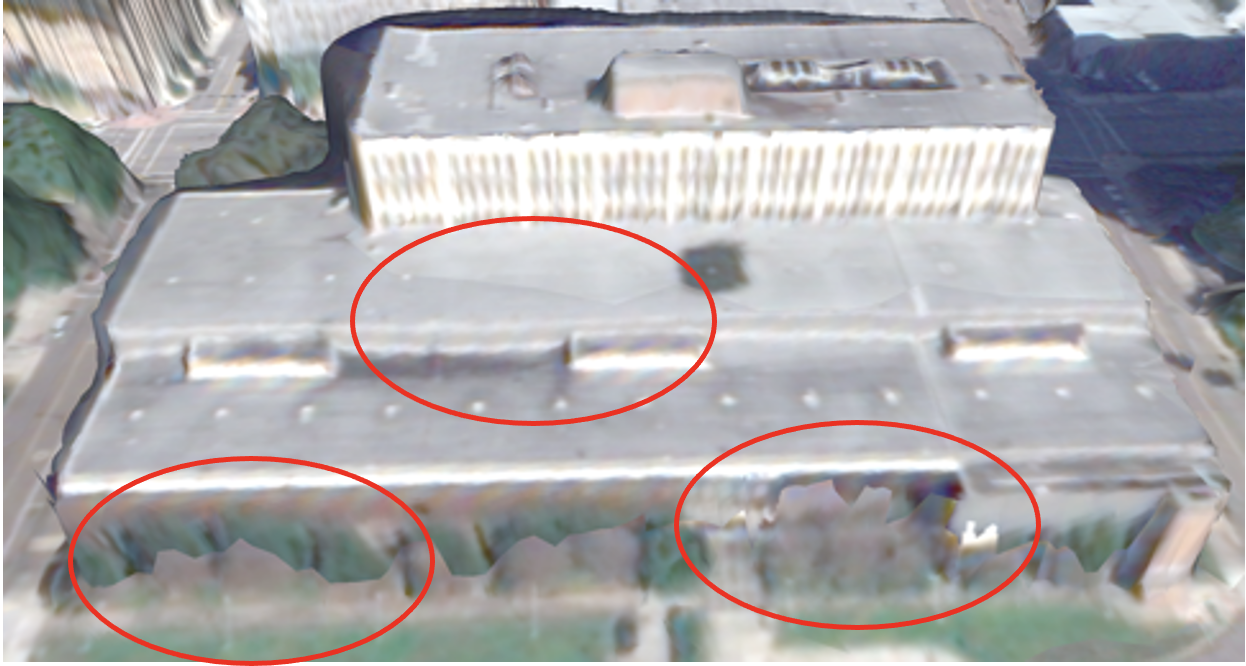}\enspace&  \includegraphics[width=0.49\columnwidth]{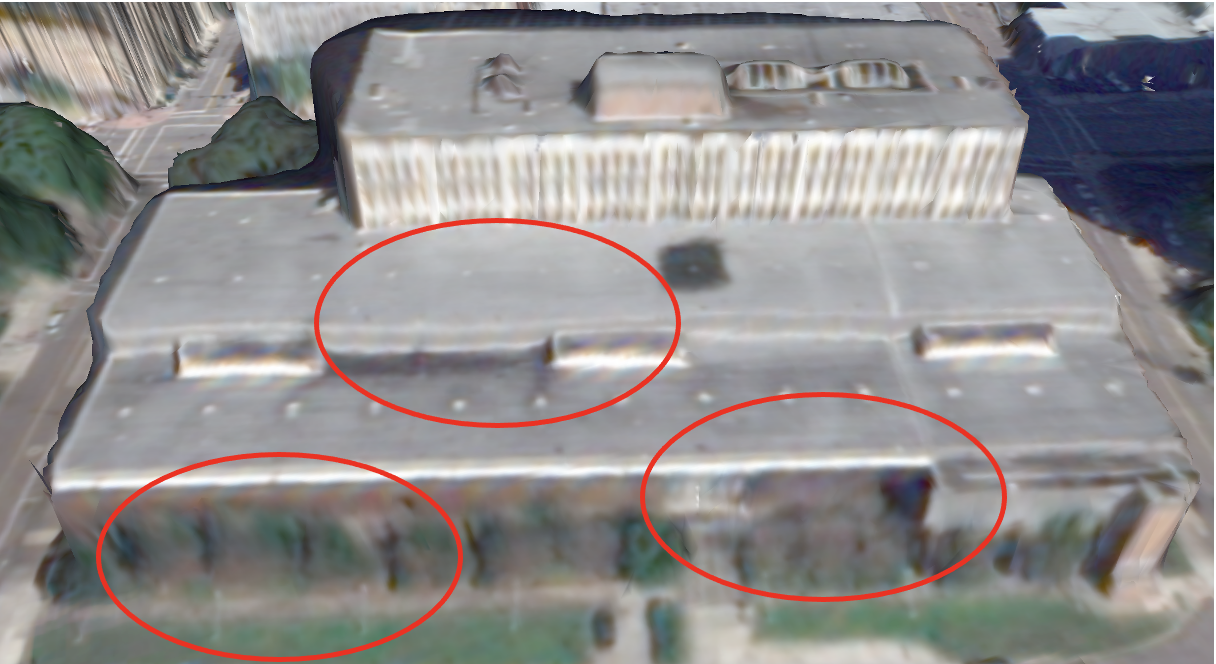}\\
    (b) & (c)
  \end{tabular}
  \caption{{Example depicting our view-blending approach. (a) first row: the
      masks for blending (highlighted in red); second row: their corresponding
      distance transformations, a higher value means a larger distance to
      the borders of the mask in the first row. (b) is a building with faces
      textured by merely the best view, similar to those with the graph-cut
      solution, while (c) is the view-blending result of the top three
      images. Regions highlighted in circles show significant improvements over
      seamlines between faces.
    }}\label{fig:view_blending}
\end{figure}

\noindent There are multiple \enquote{good} views for each mesh face after the
view selection procedure and there will still exist minor color {differences} among
them. Direct fusion with equal weights for all involved views will cause color
discontinuities and blurriness, especially for the faces whose top N views are
different. We therefore introduce a modified alpha image blending algorithm
based on distance transformation, as it was shown to be effective in image
stitching applications. The algorithm consists of the following steps: Each face
is assigned with N views and for each view, a mask consisting of all valid
pixels (corresponding to the face and non-occluded) is firstly generated for the
view {(see examples of the mask highlighted in red in the first row of \textbf{Fig.
  \ref{fig:view_blending} (a)})}, followed by computing $L^2$ distance transform
{(see the second row of \textbf{Fig. \ref{fig:view_blending} (a)})}, the final texture for each face
is obtained by blending all assigned view data together with associated distance
transforms as weights {(see \textbf{Fig. \ref{fig:view_blending} (c)})}. If
assigning the single and best view to map the texture{, as shown in \textbf{Fig.
  \ref{fig:view_blending} (b)},} it will show apparent luminance differences,
which can be greatly reduced by the proposed view blending approach {(see
\textbf{Fig. \ref{fig:view_blending} (c)})}.

\section{Experiments}\label{sec:experiment}
\noindent Our proposed method has been tested on three different types of
datasets (Satellite dataset, unmanned-aerial vehicle (UAV) dataset, and
close-range dataset) of varying complexity{. Specific} data parameters are
listed in \textbf{Table \ref{tab:datasets}}. All datasets are tested on a
workstation with two 14-core Xeon W-2275 CPUs and 128 GB of memory. To
demonstrate the effectiveness and efficiency of the proposed method, the
proposed approach is variably compared against three {SOTAs}
including both blending-based and MRF-based methods: (1) masked photo blending
\cite{callieri_masked_2008}, which generates multiple masks per image, e.g.
intersection angle, depth map border map, etc., then combine all masks into maps
as weights for images in the final multi-view blending to yield the mesh
texture. It is a classic blending-based TM method, which is simple and
applicable for different types of camera models, used as {a} baseline in our
experiments; (2) MVS (Multi-View Stereo)-based texture mapping
\cite{waechter_let_2014} that builds on the MRF energy function to select an
optimal view for each mesh face, followed by global color adjustment and Poisson
editing \cite{perez_poisson_2003} to minimize visible luminance difference
between patches. Since it does not support RPC as {a} camera model and is
designed for large-scale texturing, we use it as {a} comparison in the UAV dataset;
(3) patch-based optimization for TM \cite{bi_patch-based_2017}, which
synthesizes aligned images from input images even with inaccurate camera poses
by proposing a patch-based optimization system to avoid ghosting and blurring
artifacts in the result. However, this method is not scalable (see
\textbf{Section \ref{sec:expr_close_range}}), {and thus it} is only used in
the close-range dataset. In addition, a well-known commercial photogrammetry
software Metashape Pro \cite{agisoft_llc_metashape_2021} is also used in all
three experiments to generate the mesh textures from input 3D mesh models and
image data as {a} comparison, the algorithm of which is unknown.

\begin{table*}[t]
  \centering
  \begin{threeparttable}
  \caption{Summary of the three datasets.}
  \begin{tabular}{cccccc}
    \hline
    \hline
    dataset&\#views&\#mesh faces&source image dimension&orientation parameters&compared method\\
    \hline
    Satellite&12&1.9 million&$42253\times 32125$&RPC parameter after bundle adjustment&(1),(4)\\
    UAV&113&2.4 million&$5472\times 3648$&Calibrated camera poses&(1),(2),(4)\\
    Close-range&33&0.06 million&$640\times 480$&Camera poses&(1),(3),(4)\\
    \hline
    \hline
  \end{tabular}
  \begin{tablenotes}
    \small
  \item (1) Callieri et al.’s method \cite{callieri_masked_2008}; (2) Waechter et al.‘s method \cite{waechter_let_2014}; (3) Bi et al.’s method \cite{bi_patch-based_2017}; (4) Metashape Pro \cite{agisoft_llc_metashape_2021}.
  \end{tablenotes}
  \end{threeparttable}
  \label{tab:datasets}
\end{table*}

We implemented our framework and Callieri et al.’s method
\cite{callieri_masked_2008} in C/C++ and used the
authors’ code for Waechter et al. \cite{waechter_let_2014}. For Bi et al.’s approach\cite{bi_patch-based_2017}, an available
open-source implementation on GitHub \cite{eagle_eagle-texturemapping_2021} was
used. We used default parameter configurations for all these compared approaches
since these were tested to perform the best empirically. Specifically, in the
MRF energy optimization, although as mentioned in {Lempitsky and Ivanov's
  work \cite{lempitsky_seamless_2007}}, the GC solver (i.e. $\alpha$-expansion) achieves the best performance over the LBP solver, however, unlike the GC solver which cannot be parallelized easily due to the graph’s irregular structure, the LBP solver has been fully parallelized in our implementation, which yields greater resource use and computational efficiency. 

The evaluations of these three datasets are performed in the following three
subsections, where we performed both 1) qualitative evaluation (visual
comparison), 2) quantitative evaluation, and 3) {running time} and memory
use. {Most of the SOTAs evaluate their methods through
  either visual assessment or running time.
As far as we know, there is no generally accepted method or indicator to
measure the accuracy and reliability of the 3D texture. Thus, we designed a new texture quality
assessment method, following the intuition that the rendered image from the
textured model should coincide with the original image. Then, the difference
between the rendered and the original image is assessed using the Peak
signal-to-noise ratio (PSNR) and the multiscale structural similarity index
measure (MS-SSIM) \cite{wang_multiscale_2003}.
}
More specifically, for an input image $I_i$ and its associated camera pose (or
RPC for satellite data) $pos_i$, a virtual image $V_i$ with the same resolution
as $I_i$ {was} firstly captured at $pos_i$ from one textured result, and the
image quality for generated $V_i$ was then quantified in terms of some selected
indicators (e.g. PSNR and MS-SSIM in our experiment) with respect to the
original image $I_i$, these assessments were collected for all input images, the
mean PSNRs and the mean MS-SSIMs were further computed over all assessments as
quality indicators for the whole dataset. Note that, for both PSNR and MS-SSIM,
{the higher the value, the higher the image quality}.

\subsection{Experiment on the satellite dataset} \label{sec:expr_satellite}
\begin{figure}[ht]
  \centering
  \begin{tabular}{@{}c@{}c}
    \includegraphics[width=0.4\columnwidth,height=0.32\columnwidth]{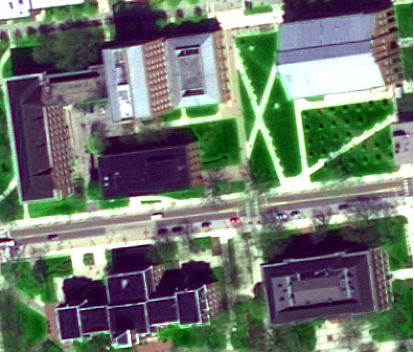}\enspace&  \includegraphics[width=0.4\columnwidth,height=0.32\columnwidth]{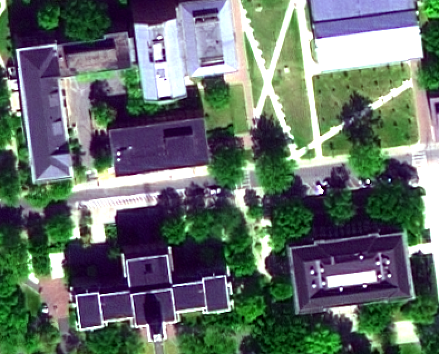}\\
    (a)&(b)\\
  \end{tabular}
  \caption{{Two satellite images with different viewing angles. (a)
      shows an off-nadir image with visible fa\c{c}ades; (b) shows primarily
      textures of roofs in a nearly nadir view.}}\label{fig:expr_satellite_viewing}
\end{figure}

\noindent
Twelve 0.5m GSD (Ground Sampling Distance) WorldView I/II satellite images
covering the main campus of the Ohio State University (OSU) in Columbus, Ohio,
USA, are collected and processed using the satellite stereo processing software
RSP\cite{qin_rpc_2016} to generate the plain mesh for this region.
{Due to the large coverage and the high angular agility of the WorldView I/II
  satellites, all satellite images in the dataset cover the whole region at
  various angles (up to 15 degrees of off-nadir angle). An example is shown in
  \textbf{Fig. \ref{fig:expr_satellite_viewing}}, where a near nadir image
  contains minimal distortion on the roofs, but does not contain visible
  textures of the fa\c{c}ades, while an off-nadir image contains visible
  fa\c{c}ades with less optimal roof textures. Therefore, an independent
  per-face quality indicator will likely select \textbf{Fig.
    \ref{fig:expr_satellite_viewing}(a)} for fa\c{c}ade textures and \textbf{Fig.
    \ref{fig:expr_satellite_viewing}(b)} for roofs, which inevitably causes seamlines.
  Our proposed approach, instead of selecting one of them, will take both and
  blend them to explore the advantages of both images, at the same time, create
  seamless texture mapping results across roof and fa\c{c}ade boundaries.
}

\begin{figure}[ht]
  \centering
  \includegraphics[width=\columnwidth]{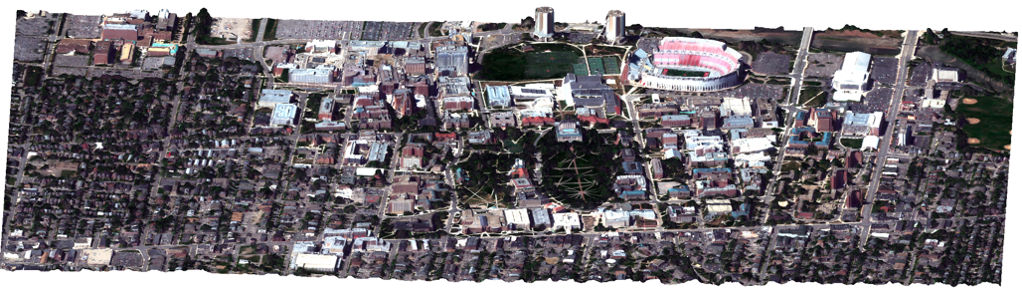}\\
  \begin{tabular}{@{}c@{}c}
    \includegraphics[width=0.49\columnwidth]{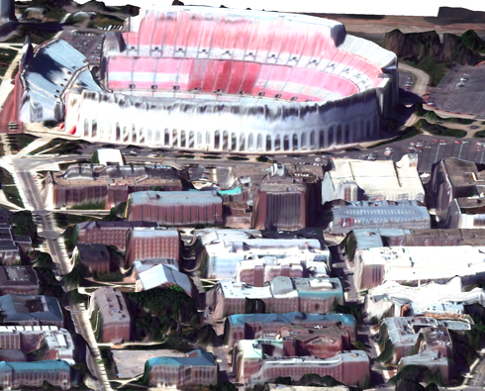}\enspace&  \includegraphics[width=0.49\columnwidth]{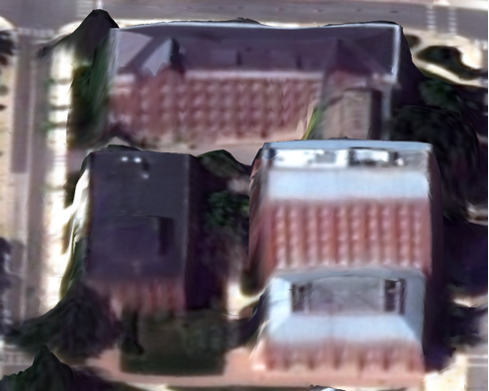}\\
  \end{tabular}
  \caption{Examples of texture mapping results of the satellite dataset. Top: the overview of the textured mesh. Bottom: enlarged views to show details on fa\c{c}ades.}\label{fig:expr_satellite_overview}
\end{figure}

TM for this region using the proposed method is completed within 4 minutes and consumed 4 GB peak memory, yielding a visually consistent textured mesh as shown in \textbf{Fig. \ref{fig:expr_satellite_overview}}. Two other methods are used as a comparison: 1) Metashape Pro\cite{agisoft_llc_metashape_2021} and 2) Callieri et al.’s method\cite{callieri_masked_2008}.

\begin{figure}[ht]
  \centering
  \begin{tabular}{@{}c@{}c}
    \includegraphics[width=0.45\columnwidth,height=0.45\columnwidth]{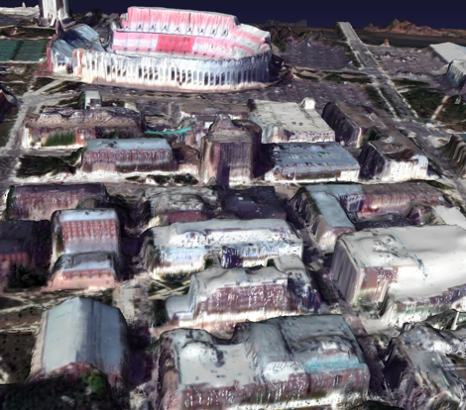}\enspace&  \includegraphics[width=0.45\columnwidth,height=0.45\columnwidth]{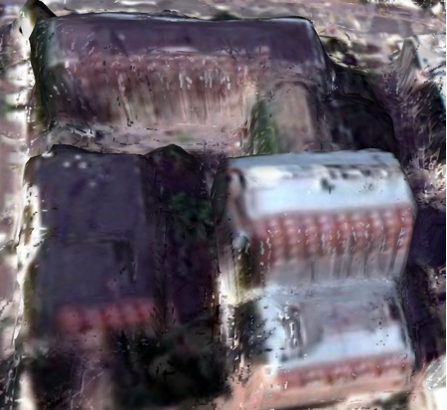}\\
  \end{tabular}
  \caption{Some texturing results from Metashape Pro. White noises can be seen everywhere.}\label{fig:expr_satellite_metashape}
\end{figure}

Metashape Pro with its newest support for satellite datasets was applied to this
dataset and the results are shown in \textbf{Fig.
  \ref{fig:expr_satellite_metashape}}. We observed that the textured mesh
generated with default configurations reflected a large number of systematic
errors. Since no warnings or errors were found in its log file, the authors
perceive the results are not numerically comparable {and} potentially due to
unknown system errors. {Thus, it} is not included for further qualitative comparison. 

The comparisons {of running time} \& memory use and quantitative evaluation are listed in \textbf{Table \ref{tab:expr_satellite_comparison}}. Callieri’s method used slightly fewer memories than ours, however, our method {is} faster, because Callieri’s method requires computations in generating various weight maps for different quality indicators compared to our well-designed efficient data term, besides, its blending function involves all input images for each mesh face, while ours only blends the selected top N views. In addition, two quantitative indicators, i.e. the mean PSNR and mean MS-SSIM over the entire test dataset, show that the textured results generated from both Callieri’s blending-based TM method and ours have comparable quality, while ours uses fewer and highly selective views. 

\begin{table}[ht]
  \centering
  \caption{{Running time} \& peak memory and quantitative evaluation for different TM methods applied to the satellite dataset.}
  \begin{tabular}{ccccc}
    \hline
    \hline
    &{Running time}&Peak memory &Mean&Mean\\
    &(sec)&usage (GB)&PSNR&MS-SSIM\\
    \hline
    Callieri et al.&270&\textbf{3.79}&16.26&0.60\\
    Ours&\textbf{232}&3.84&\textbf{16.35}&\textbf{0.61}\\
    \hline
    \hline
  \end{tabular}
  \label{tab:expr_satellite_comparison}
\end{table}

\begin{figure}[ht]
  \centering
  \small
  \begin{tabular}{@{}c@{}c@{}c@{}c}
    \includegraphics[width=0.24\columnwidth,height=0.27\columnwidth]{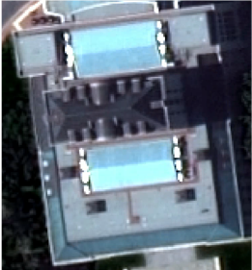}\;& \includegraphics[width=0.24\columnwidth,height=0.27\columnwidth]{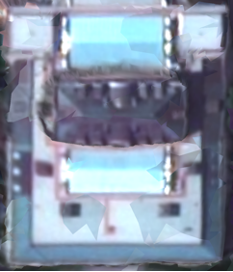}\;& \includegraphics[width=0.24\columnwidth,height=0.27\columnwidth]{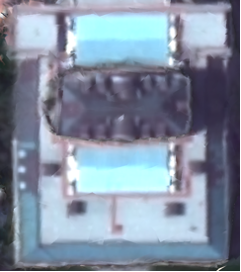}\;& \includegraphics[width=0.24\columnwidth,height=0.27\columnwidth]{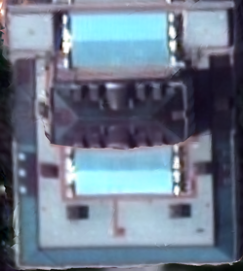}\\
    \includegraphics[width=0.24\columnwidth,height=0.27\columnwidth]{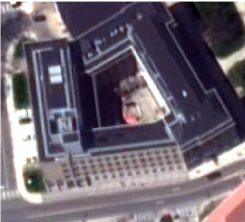}\;& \includegraphics[width=0.24\columnwidth,height=0.27\columnwidth]{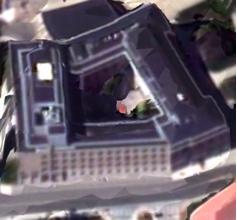}\;& \includegraphics[width=0.24\columnwidth,height=0.27\columnwidth]{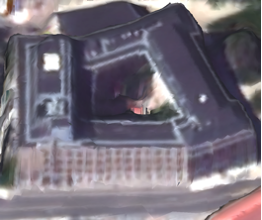}\;& \includegraphics[width=0.24\columnwidth,height=0.27\columnwidth]{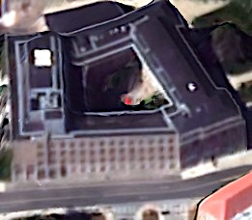}\\
    (a)&(b)&(c)&(d)\\
  \end{tabular}
  \caption{Two examples for view blending. (a) Original image; (b) rendered view using a best view selection algorithm; (c) rendered view of Callieri’s method \cite{callieri_masked_2008}; (d) ours.}\label{fig:expr_satellite_detail}
\end{figure}

Visual comparisons of different methods are shown in \textbf{Fig. \ref{fig:expr_satellite_detail}}, including 1) a best-view selection method for TM \textbf{Fig. \ref{fig:expr_satellite_detail}(b)}, 2) Callieri’s method \textbf{Fig. \ref{fig:expr_satellite_detail}(c)}, and 3) ours \textbf{Fig. \ref{fig:expr_satellite_detail}(d)}. It can be seen that the result using only the best view yields color differences among triangles, while both Callieri’s method and ours {show} consistent texture rendering. Ours \textbf{Fig. \ref{fig:expr_satellite_detail}(d)} {shows} apparently sharper texture rendering as compared to Callieri’s method, since our method uses the top-N views with optimal quality selected through the MRF framework, while Callieri’s method simply {blends} all views, which yields reduced and blurred texture. 

\subsection{Experiment on the UAV dataset}
\begin{figure}[ht]
  \centering
  \begin{tabular}{@{}c@{}c}
    \includegraphics[width=0.49\columnwidth]{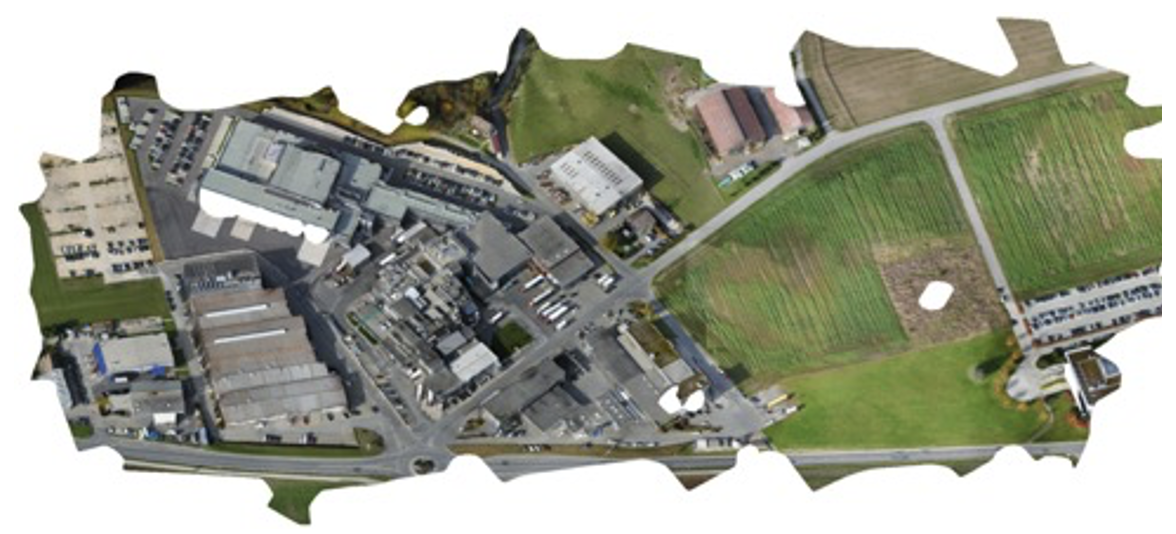}\enspace&  \includegraphics[width=0.49\columnwidth]{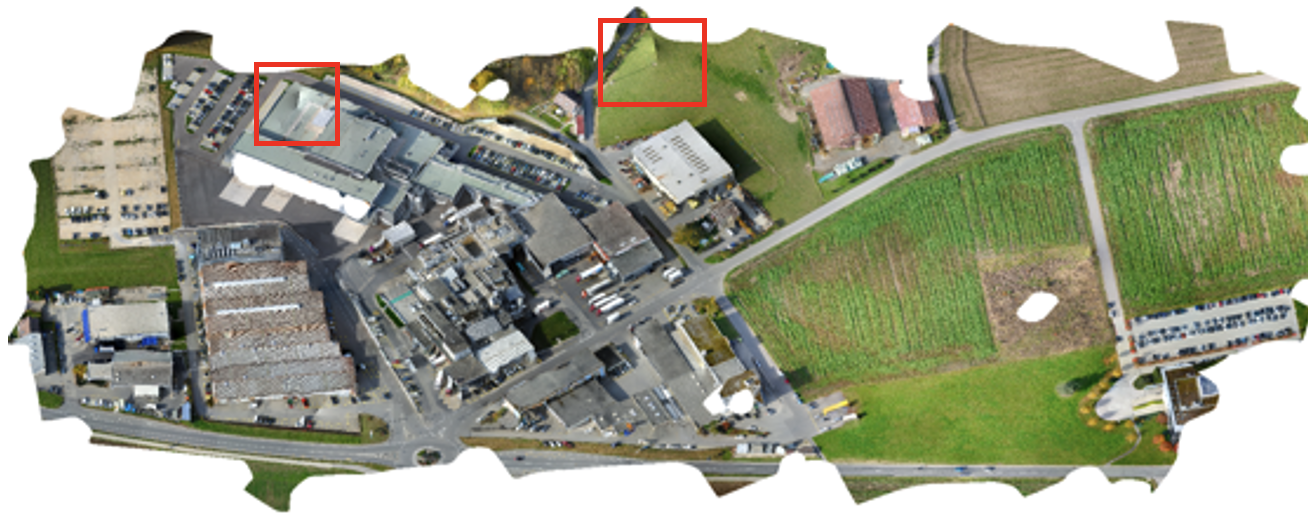}\\
    (a) Callieri &(b) Waechter\\
    \includegraphics[width=0.49\columnwidth]{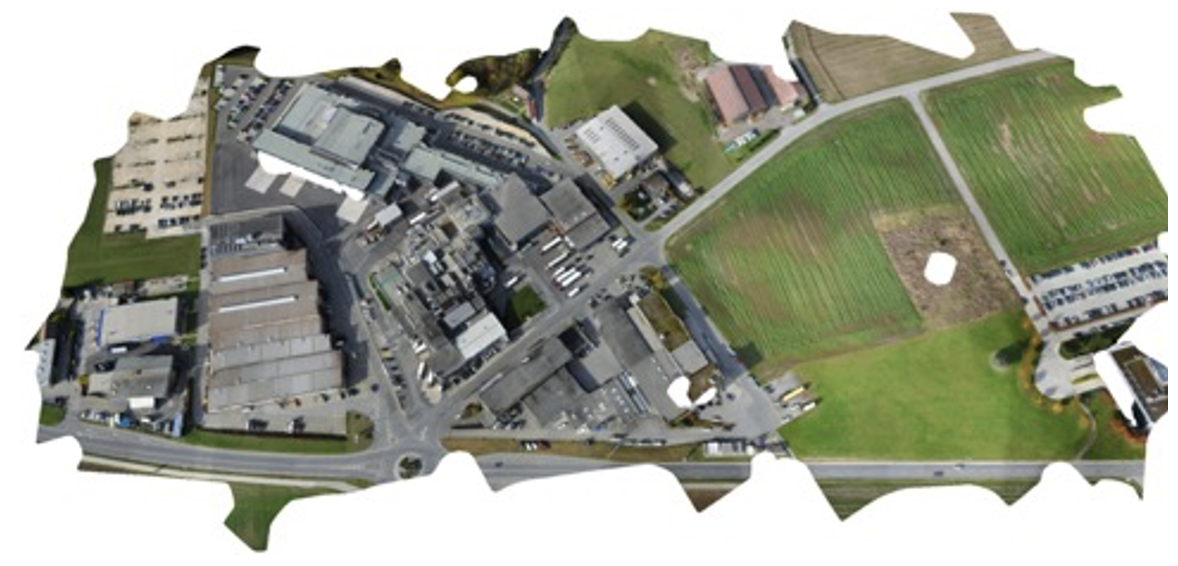}\enspace&  \includegraphics[width=0.49\columnwidth]{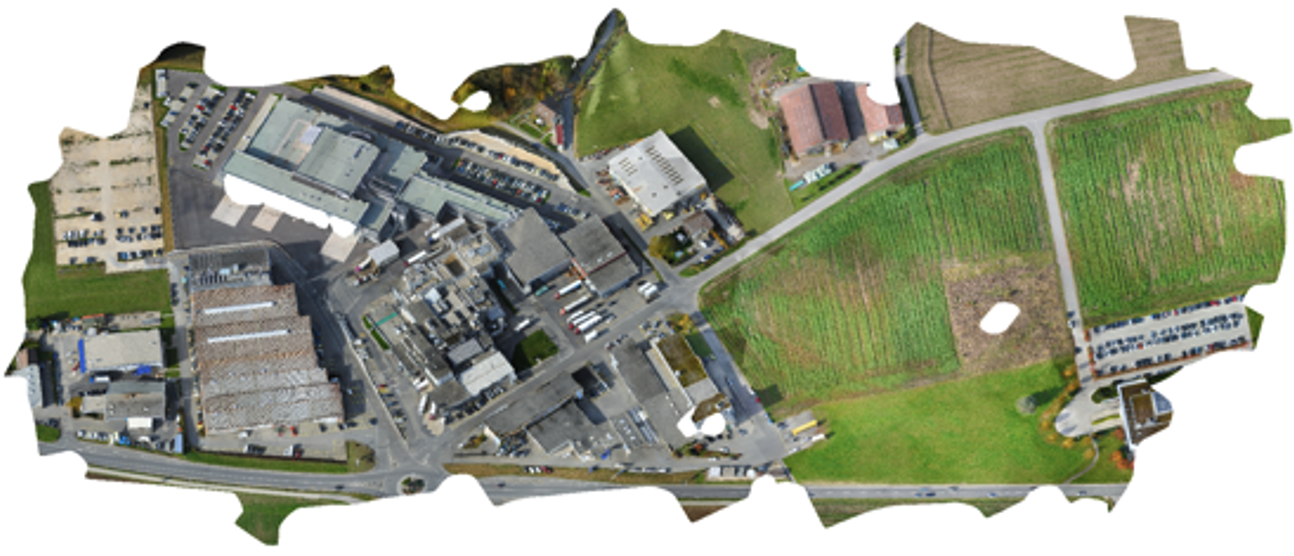}\\
    (c) Metashape Pro&(d) Ours\\
  \end{tabular}
  \caption{Overviews for four textured meshes from (a) Callieri et al.’s method \cite{callieri_masked_2008}; (b) Waechter et al.’s method \cite{waechter_let_2014}; (c) Metashape Pro \cite{agisoft_llc_metashape_2021}; (d) our proposed method. The results look similar to each other except Waechter et al’s has some artifacts highlighted in red rectangles.}\label{fig:expr_uav_overview}
\end{figure}

\noindent The UAV dataset {consists} of 113 images covering an area mixed with buildings and fields, published by senseFly \cite{sensefly_discover_2021}. We used OpenMVG \cite{moulon_openmvg_2016} to adjust all input camera poses, followed by using OpenMVS to generate the plain mesh. We applied our proposed method to produce the textured mesh compared against three {SOTAs}, i.e. 1) Callieri et al.’s\cite{callieri_masked_2008}, 2) Waechter et al’s\cite{waechter_let_2014} and 3) Metashape Pro\cite{agisoft_llc_metashape_2021}. The evaluation includes visual comparison, quantitative texture assessment and {running time} \& memory usage. The overviews for all textured results are illustrated in \textbf{Fig. \ref{fig:expr_uav_overview}}, where it can be seen that these textured models are generally of good quality. However, Waechter et al.’s introduces artifacts highlighted in red rectangles may due to its extra global seam leveling.

The {running time} \& peak memory for all methods are shown in \textbf{Table
  \ref{tab:expr_uav_comparison}} except for the Metashape Pro since it is a
black-box and interactive software. Callieri et al.’s method consumes more
computations in weight map generation and image blending (as mentioned in
\textbf{Section \ref{sec:expr_satellite}}), thus it incurs more computation
time. The total running time for Waechter et al.’s method is 20 minutes, while
ours is 16 minutes, roughly 20\% more efficient. For peak memory usage, Waechter
et al’s method is 12.82 GB while ours is 16.10 GB. Our method uses marginally
more memories mainly due to two strategies: 1) fully parallelized LBP solver
instead of GC solver in MRF energy optimization, 2) top N views are stored in
memory in preparation {for} the image blending algorithm. Quantitative evaluation in \textbf{Table \ref{tab:expr_uav_comparison}} {shows} the indicators computed from ours are slightly higher than all three other methods’ on the whole dataset, which indicates the proposed method is able to render textures closer to the input image quality than other methods.

\begin{table}[ht]
  \centering
  \caption{{Running time} \& peak memory and quantitative evaluation for different TM methods applied to the UAV dataset.}
  \begin{tabular}{ccccc}
    \hline
    \hline
    &{Running time}&Peak memory &Mean&Mean\\
    &(min)&usage (GB)&PSNR&MS-SSIM\\
    \hline
    Callieri et al. &26&14.52&16.48&0.54\\
    Waechter et al. &20&\textbf{12.82}&15.87&0.61\\
    Metashape Pro&N.A.&N.A.&16.49&0.50\\
    Ours&\textbf{16}&16.10&\textbf{16.76}&\textbf{0.64}\\
    \hline
    \hline
  \end{tabular}
  \label{tab:expr_uav_comparison}
\end{table}

\begin{figure}[ht]
  \centering
  \footnotesize
  \begin{tabular}{@{}c@{}c@{}c@{}c}
    \rotatebox{90}{\;\;\; Callieri}\;&\includegraphics[width=0.3\columnwidth,height=0.18\columnwidth]{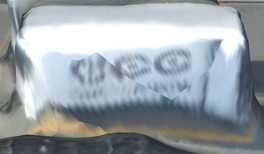}\;& \includegraphics[width=0.3\columnwidth,height=0.18\columnwidth]{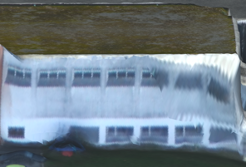}\;& \includegraphics[width=0.3\columnwidth,height=0.18\columnwidth]{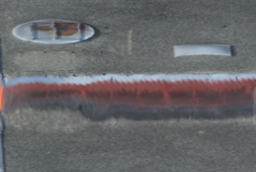}\\
    \rotatebox{90}{\;\; Waechter}\;&\includegraphics[width=0.3\columnwidth,height=0.18\columnwidth]{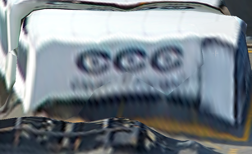}\;& \includegraphics[width=0.3\columnwidth,height=0.18\columnwidth]{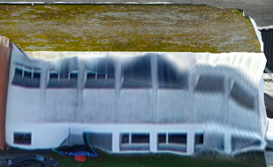}\;& \includegraphics[width=0.3\columnwidth,height=0.18\columnwidth]{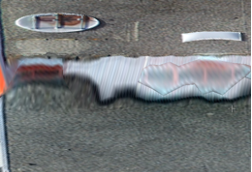}\\
    \rotatebox{90}{\scriptsize Metashape Pro}\;&\includegraphics[width=0.3\columnwidth,height=0.18\columnwidth]{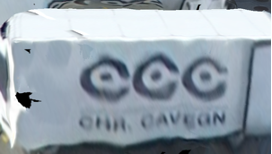}\;& \includegraphics[width=0.3\columnwidth,height=0.18\columnwidth]{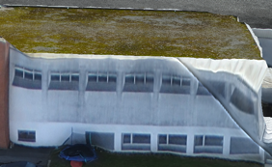}\;& \includegraphics[width=0.3\columnwidth,height=0.18\columnwidth]{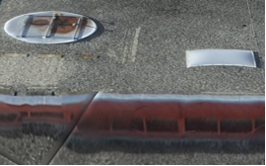}\\
    \rotatebox{90}{\;\;\;\; Ours}\;&\includegraphics[width=0.3\columnwidth,height=0.18\columnwidth]{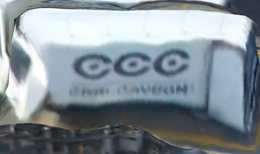}\;& \includegraphics[width=0.3\columnwidth,height=0.18\columnwidth]{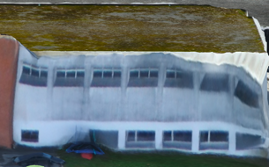}\;& \includegraphics[width=0.3\columnwidth,height=0.18\columnwidth]{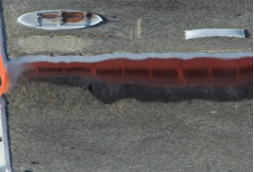}\\
  \end{tabular}
  \caption{Three examples of difficult texturing regions showing our advantages over other three methods: Callieri et al.’s method \cite{callieri_masked_2008}, Waechter et al.’s method \cite{waechter_let_2014}, Metashape Pro \cite{agisoft_llc_metashape_2021}. First row: visible seamline and noises; Second row: poor texturing faces and over color adjustment; Third row: carving artifacts; Last row: seamless high-quality texture.}\label{fig:expr_uav_detail}
\end{figure}

\textbf{Fig. \ref{fig:expr_uav_detail}} gives three different examples of
enlarged views where three {SOTAs} all produced poor texturing
results, e.g. visible seamlines and noises, tearing and discoloration. Some
issues (i.e. seamline and tearing) are caused by the corresponding blending
algorithm. For example, Callieri et al.’s method blends all candidate images, in
which poor images saturated \enquote{good} ones, resulting in blurry and
inconsistent textures (first row of \textbf{Fig. \ref{fig:expr_uav_detail}}.).
{Areas} with too few \enquote{good} views are also problematic, as the
best-view selection algorithm might not accurately select the real
\enquote{best} one. For example, we observe from {the} result of Waechter et
al.’s method (second row of \textbf{Fig. \ref{fig:expr_uav_detail}}.), the
fa\c{c}ade textures are not necessarily {of} the best quality, as it was
geared towards {having} the same view as neighboring faces, which in this
case are roof faces, thus using the same view as used by the roof faces result
in the loss of resolution. In addition, although no technical details about its
TM algorithm are known for Metashape Pro, it yielded a high-quality texture
except {for} some apparent artifacts {that may be} due to over color adjustment (third row of \textbf{Fig. \ref{fig:expr_uav_detail}}.). However, our proposed method is less problematic {for} all these issues, as it leverages \enquote{good} views not only on each face, but also across neighboring faces during weighted blending (last row of \textbf{Fig. \ref{fig:expr_uav_detail}}.). 

\begin{figure}[ht]
  \centering
  \footnotesize
  \begin{tabular}{@{}c@{}c@{}c@{}c}
    \includegraphics[width=0.24\columnwidth,height=0.16\columnwidth]{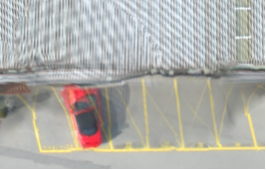}\;& \includegraphics[width=0.24\columnwidth,height=0.16\columnwidth]{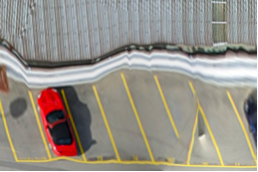}\;& \includegraphics[width=0.24\columnwidth,height=0.16\columnwidth]{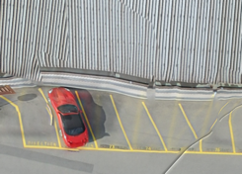}\;& \includegraphics[width=0.24\columnwidth,height=0.16\columnwidth]{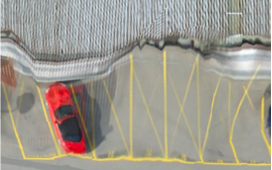}\\
    Callieri&Waechter&Metashape Pro&Ours\\
  \end{tabular}
  \caption{An example showing drawbacks in the proposed methods at its ghosting artifacts due to inaccurate geometry/camera poses.}\label{fig:expr_uav_drawback}
\end{figure}

Since our proposed method in nature blends multiple views, it is still inherently affected by inaccuracies of either the geometry or camera poses, as shown in \textbf{Fig. \ref{fig:expr_uav_drawback}}, where these single-view methods apparently achieved better results, and although our result shows ghosting effects, it still outperforms the other blending-based method (e.g. Callieri et al. ).

In conclusion, our proposed method can generate the mesh texture in {slightly} better quality than other {SOTAs}, and consume less computation (i.e. 20\% faster), overcome several artifacts found in Callieri’s, Waechter’s and Metashape Pro. However, ghosting artifacts is a remaining challenge in our future work.

\subsection{Experiment on the close-range dataset}\label{sec:expr_close_range}
\begin{figure}[ht]
  \centering
  \footnotesize
  \begin{tabular}{@{}c@{}c@{}c@{}c}
    \includegraphics[width=0.35\columnwidth,height=0.35\columnwidth]{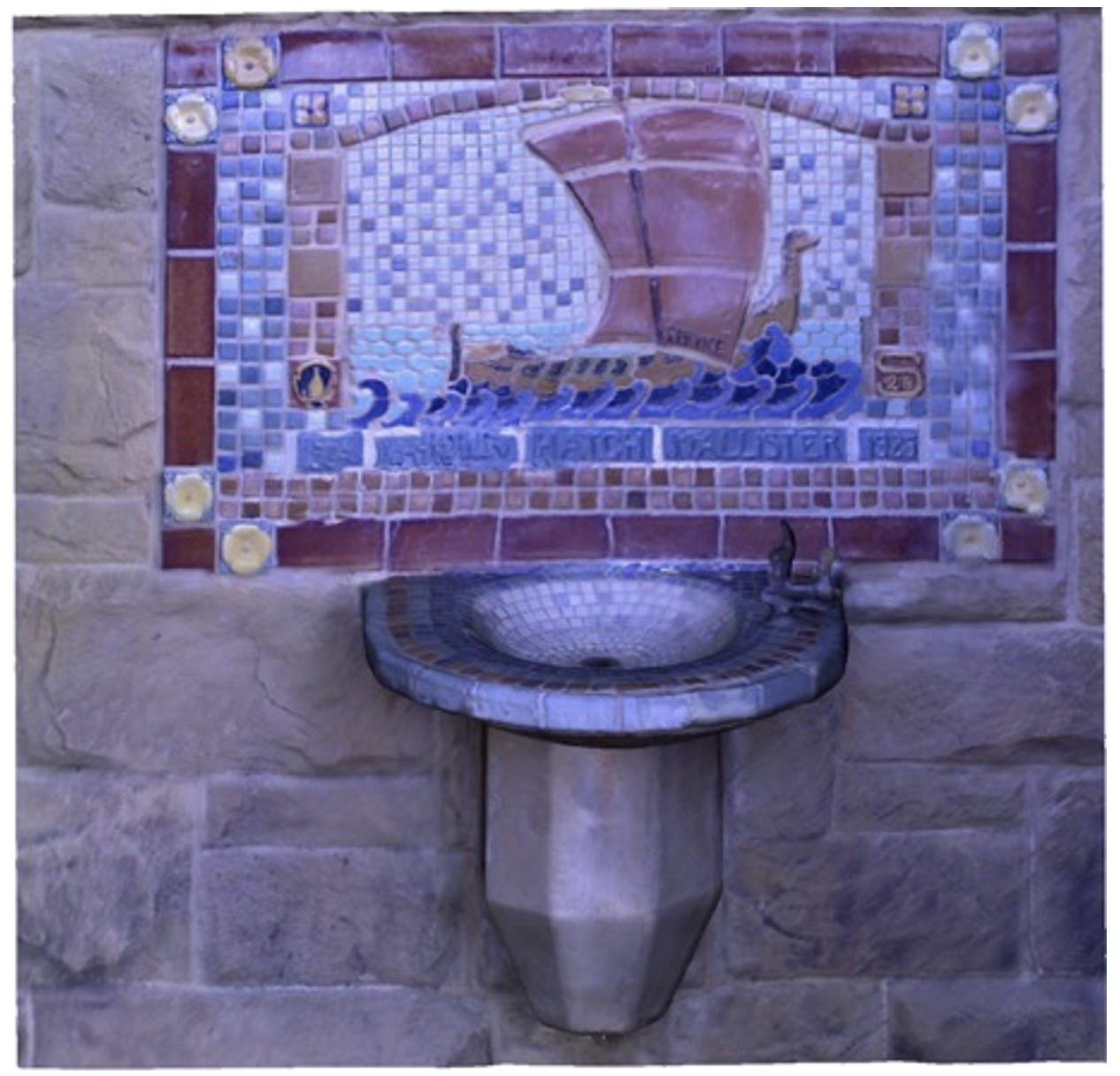}\;\;& \includegraphics[width=0.35\columnwidth,height=0.35\columnwidth]{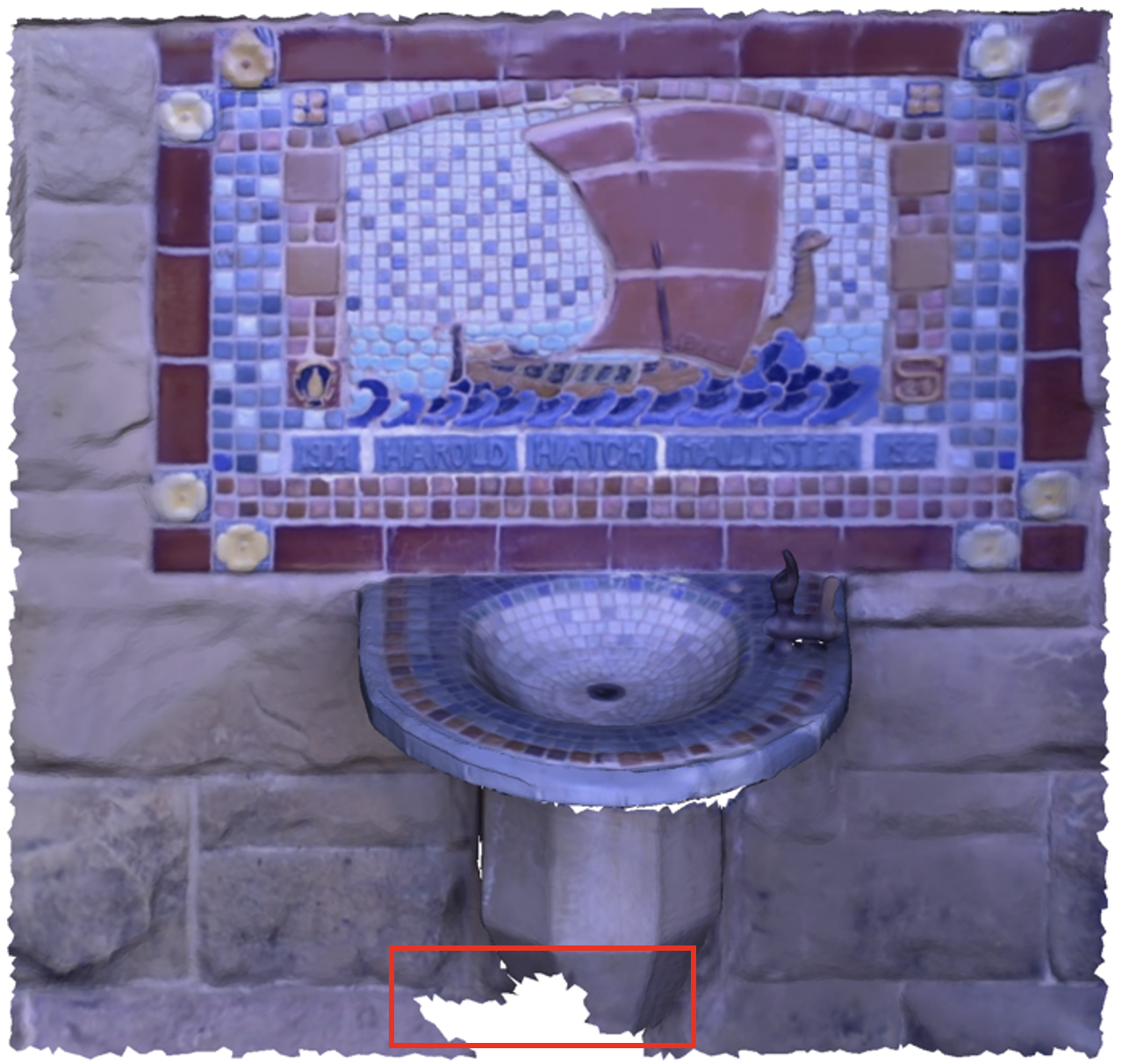}\\
    (a) Callieri&(b) Bi\\
    \includegraphics[width=0.35\columnwidth,height=0.35\columnwidth]{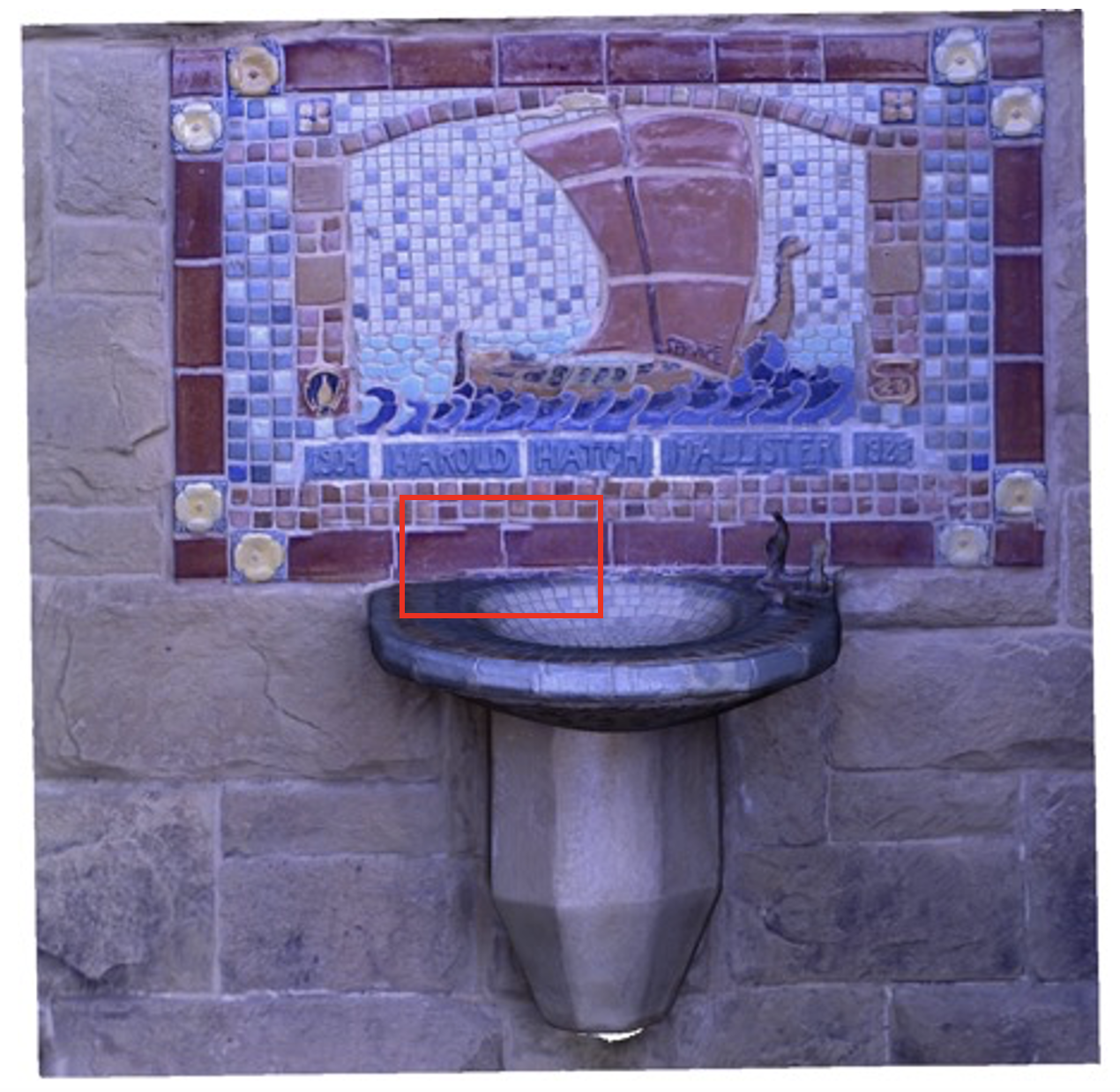}\;\;& \includegraphics[width=0.35\columnwidth,height=0.35\columnwidth]{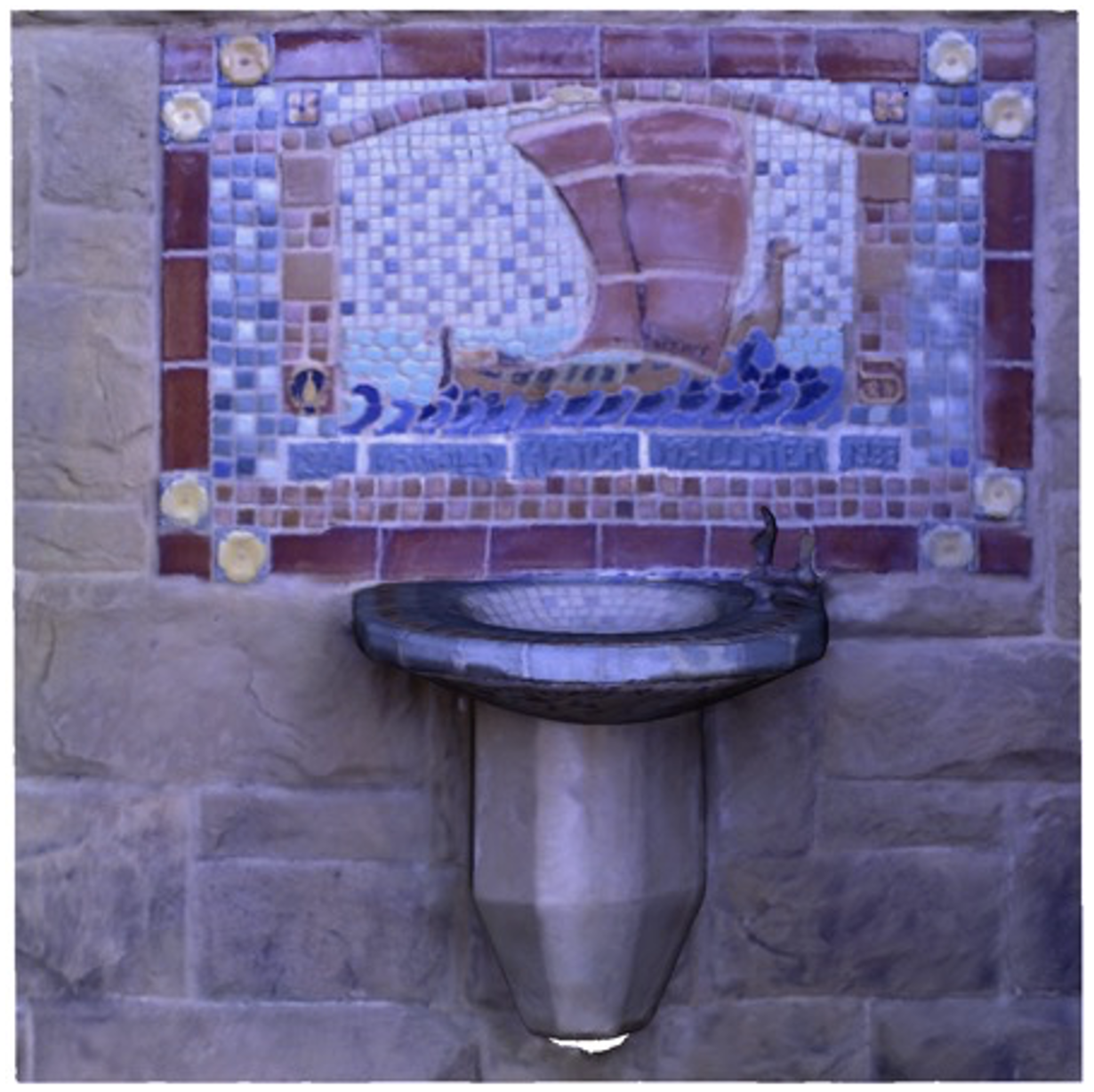}\\
    (c) Metashape Pro&(d) Ours\\
  \end{tabular}
  \caption{Overviews for four textured meshes from (a) Callieri et al.’s method
    \cite{callieri_masked_2008}; (b) Bi et al.’s method
    \cite{bi_patch-based_2017}; (c) Metashape Pro
    \cite{agisoft_llc_metashape_2021}; (d) our proposed method. Callieri’s and
    ours are similar to each other and are both blurrier than {the} other two methods (see the texts in the middle of pictures). However, there is apparently tearing in Metashape Pro’s result as highlighted in rectangle and Bi’s result, although with {fewer} artifacts, has missing parts near borders.}\label{fig:expr_cr_overview}
\end{figure}

\noindent The Fountain scene from Zhou and Koltun \cite{zhou_color_2014} is used
to discuss the limitations of our proposed method. Since the plain mesh was
already provided along with 33 images, we directly tested the proposed method
against three {SOTAs}, i.e. 1) Callieri’s method\cite{callieri_masked_2008}, 2) Bi’s method\cite{bi_patch-based_2017} and 3)
Metashape Pro\cite{agisoft_llc_metashape_2021}, in three aspects: (1)
{running time} \& peak memory usage; (2) quantitative evaluation and (3)
visual check. The overviews for all textured results are illustrated in
\textbf{Fig. \ref{fig:expr_cr_overview}}, where we can see all four methods have
their own limitations, such as Callieri’s and ours are blurrier than {the} other two’s, especially for the texts in the middle. Bi’s has missing parts near borders because the number of visible views for that regions don’t meet its requirement, and Metashape Pro’s has apparently tearing as highlighted in \textbf{Fig. \ref{fig:expr_cr_overview}(c)}. 

As {a} performance comparison listed in \textbf{Table \ref{tab:expr_cr_comparison}}, it costs only 24 seconds and 193 MB memories for the proposed method to generate the textured mesh, faster than Callieri’s method, the same as in previous experiments, while Bi’s method consumed 120 MB memories and took over 1 hour to generate results, two orders of magnitude slower than ours, proven to be not-applicable for TM of large dataset. \textbf{Table \ref{tab:expr_cr_comparison}} shows that all the quantitative measures are lower than those in the previous experiments, which implies the challenge of this dataset, which may be due to inaccurate input camera poses and strong illumination and exposure differences exhibited in the input images. Our method is the fastest among the compared methods and achieves the second best result in texture measure (only marginally worse than the Metashape Pro). 

\begin{table}[ht]
  \centering
  \caption{{Running time} \& peak memory and quantitative evaluation for different TM methods applied to the close-range dataset.}
  \begin{tabular}{ccccc}
    \hline
    \hline
    &{Running time}&Peak memory &Mean&Mean\\
    &(sec)&usage (MB)&PSNR&MS-SSIM\\
    \hline
    Callieri et al. &39&207&12.42&0.31\\
    Bi et al. & $>$3600&\textbf{120}&12.76&0.38\\
    Metashape Pro&N.A.&N.A.&\textbf{12.84}&\textbf{0.39}\\
    Ours&\textbf{24}&193&12.78&0.38\\
    \hline
    \hline
  \end{tabular}
  \label{tab:expr_cr_comparison}
\end{table}

\begin{figure}[ht]
  \centering
  \footnotesize
  \begin{tabular}{@{}c@{}c@{}c@{}c}
    \includegraphics[width=0.24\columnwidth,height=0.24\columnwidth]{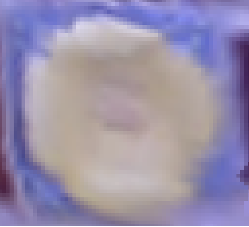}\;& \includegraphics[width=0.24\columnwidth,height=0.24\columnwidth]{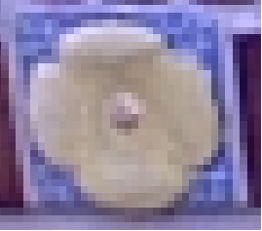}\;&\includegraphics[width=0.24\columnwidth,height=0.24\columnwidth]{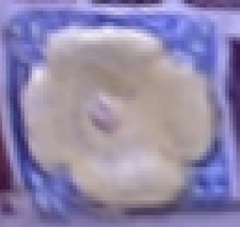}\;&\includegraphics[width=0.24\columnwidth,height=0.24\columnwidth]{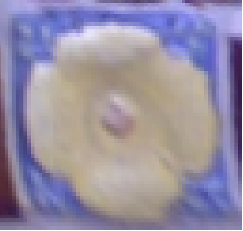}\\
    \includegraphics[width=0.24\columnwidth,height=0.24\columnwidth]{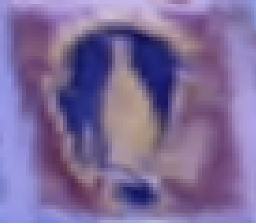}\;& \includegraphics[width=0.24\columnwidth,height=0.24\columnwidth]{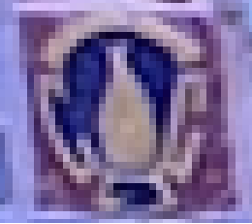}\;&\includegraphics[width=0.24\columnwidth,height=0.24\columnwidth]{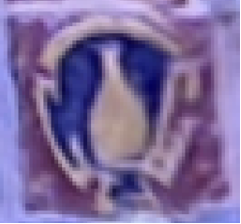}\;&\includegraphics[width=0.24\columnwidth,height=0.24\columnwidth]{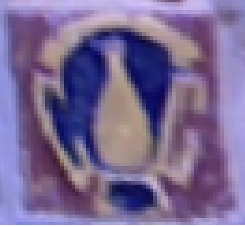}\\
    \includegraphics[width=0.24\columnwidth,height=0.24\columnwidth]{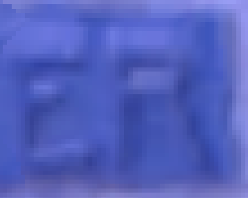}\;&\includegraphics[width=0.24\columnwidth,height=0.24\columnwidth]{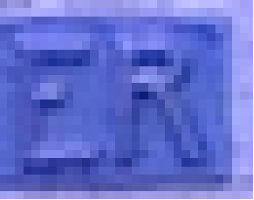}\;&\includegraphics[width=0.24\columnwidth,height=0.24\columnwidth]{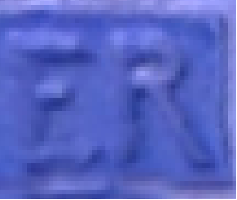}\;&\includegraphics[width=0.24\columnwidth,height=0.24\columnwidth]{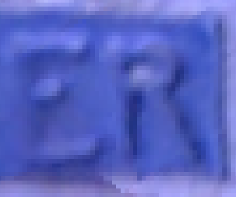}\\
    (a) Callieri&(b) Bi&(c) Metashape Pro&(d) Ours\\
  \end{tabular}
  \caption{Visual comparison for four texture mapping results on the close-range
    dataset: (a) Callieri et al.’s result \cite{callieri_masked_2008}; (b) Bi et
    al.’s result \cite{bi_patch-based_2017}; (c) Metashape Pro’s result
    \cite{agisoft_llc_metashape_2021}; and (d) ours. Ours is blurrier than Bi et
    al.’s and Metashape Pro’s due to multi-view blending with inaccurate camera
    poses, but is better than Callieri et al.’s. For Bi et al.’s, although no
    blurring or ghosting artifacts are found, seems to have {a} lower
    resolution than Metashape Pro’s, which has {the} highest quality texture with more details.}\label{fig:expr_cr_detail}
\end{figure}

We note that these inherent challenges, especially the inaccurate camera poses,
lead to slight blurring and ghosting artifacts, even distortions in the final
texture for our proposed method (see \textbf{Fig. \ref{fig:expr_cr_detail}}).
However, when checking details against {the} other three results, although
ours has a little more distortions and {is} slightly blurrier than Metashape Pro’s
and Bi’s, our result is overall reasonable and acceptable, especially considering
its efficiency and capability to process other types of datasets as well. 

\section{Conclusion}\label{sec:conclusion}
\noindent
{Typical texture mapping algorithms for aerial photogrammetry applications
  are designed to process regular images collected under consistent lighting
  conditions. However, this presents a challenge when these algorithms are
  scaled to process multi-view and large-frame images, where a large number of
  large-sized images under drastically different lighting conditions brings about
a highly non-linear image intensity difference and order-of-magnitude number of
faces to optimize. In this paper, we present a novel texture mapping framework
for large-scale 3D reconstructions, which are able to achieve better accuracy as
well as efficiency than other SOTAs across both space-borne and aerial-borne
platforms, due to: }
  \begin{enumerate}
    \item \textbf{Top-N best views instead of \enquote{one-best-view}.} Typical
      MRF-based methods find the \enquote{one-best-view}, which may
      select suboptimal views for mesh faces without chances to correct, i.e. it
      is impossible to leverage texture fusion algorithms once it is set. This
      is typically problematic for high-altitude images. When selecting views
      for faces connecting the roof and the fa\c{c}ade of a building, such algorithms
      would often select a single suboptimal view for the fa\c{c}ade, in order to
      favor spatial smoothness constraints, and result in distorted textures.
      However, ours produces the top-N best views per face, giving an additional
      chance to perform post-patch-fusion for conflicting view selections. In
      this case, ours can achieve high-quality textures for both roof and
      fa\c{c}ade faces, as evidenced by \textbf{Section \ref{sec:experiment}-A}.
    \item \textbf{More general data term.} We provide a more general data term
      designated to support more objectives and generic enough to support
      multiple camera models (e.g. camera poses and RPCs as well). This
      provides an important feature when using both satellite images and
      aerial/UAV images to generate texture from combined sources.
    \item \textbf{Simple and efficient image blending algorithm}. We
      discard the typical complex post color correction scheme, and instead
      use a local blending scheme to minimize seamlines. This is not
      regarded as a compromise, but a benefit from the selection of N
      views instead of one. It provides much more flexibility for an
      efficient image blending algorithm to achieve color consistency
      without additional color correction, and preserves the original input
      texture to the greatest extent as well.
    \item \textbf{Fully parallelized LBP solver}. In the MRF energy
      optimization, the GC solver is the most widely used solver due to its
      best performance over other solvers on a single CPU core. However, unlike the GC solver
      which cannot be parallelized easily due to the graph’s irregular
      structure, the LBP solver has been fully parallelized in our
      implementation, which yields greater resource use and computational
      efficiency.
  \end{enumerate}



The proposed method is
evaluated with three typical datasets: {a} wide-area satellite dataset (covering 4
$km^2$), {a} UAV dataset containing over 100 images and {a} close-range dataset named
\enquote{Fountain}. We compared our proposed method with four {SOTAs}
 through (1) visual quality check; (2) quantitative assessment; (3)
{running time} performance evaluation. Experimental results show that the
proposed method achieves textured meshes with more consistent and
high-resolution texture, even in very difficult scenarios such as faces connecting
fa\c{c}ades and roofs. {At the meantime, it is} at least 20\% faster than all the other methods
with only a slightly increased memory use. Moreover, among all these tested
methods, our method consistently achieves reasonable results across the tested
dataset, and is shown as the only one {achieving} a visually pleasant result in the
satellite dataset. {Therefore, it is} flexible and scalable to handle all the tested
scenarios at different ranges of resolution and scene complexity. During our
evaluation, we also observed that, since our method relies on multiple
views, it may yield {a} slightly blurry texture for inaccurate geometry and camera
poses. Thus, our future effort will focus on adaptively determining the
number of views to use during TM to address this challenge.

\section{Acknowledgments}
\noindent This work is partially supported by Office of Naval Research (Award No. N000141712928 \& N000142012141). We thank SenseFly for making the drone dataset available and thank Debao Huang for his assistance in running part of the experiments.







\begin{thebibliography}{10}
\providecommand{\url}[1]{#1}
\csname url@samestyle\endcsname
\providecommand{\newblock}{\relax}
\providecommand{\bibinfo}[2]{#2}
\providecommand{\BIBentrySTDinterwordspacing}{\spaceskip=0pt\relax}
\providecommand{\BIBentryALTinterwordstretchfactor}{4}
\providecommand{\BIBentryALTinterwordspacing}{\spaceskip=\fontdimen2\font plus
\BIBentryALTinterwordstretchfactor\fontdimen3\font minus
  \fontdimen4\font\relax}
\providecommand{\BIBforeignlanguage}[2]{{%
\expandafter\ifx\csname l@#1\endcsname\relax
\typeout{** WARNING: IEEEtran.bst: No hyphenation pattern has been}%
\typeout{** loaded for the language `#1'. Using the pattern for}%
\typeout{** the default language instead.}%
\else
\language=\csname l@#1\endcsname
\fi
#2}}
\providecommand{\BIBdecl}{\relax}
\BIBdecl

\bibitem{gruen_operable_2020}
\BIBentryALTinterwordspacing
A.~Gruen, G.~Schrotter, S.~Schubiger, R.~Qin, B.~Xiong, C.~Xiao, J.~Li,
  X.~Ling, and S.~Yao, ``\BIBforeignlanguage{en}{An {Operable} {System} for
  {LoD3} {Model} {Generation} {Using} {Multi}-{Source} {Data} and
  {User}-{Friendly} {Interactive} {Editing}: {Final} {Report}},'' ETH Zurich,
  Report, 2020, accepted: 2020-04-14T08:54:52Z. [Online]. Available:
  \url{https://www.research-collection.ethz.ch/handle/20.500.11850/409483}
\BIBentrySTDinterwordspacing

\bibitem{hoegner_automatic_2016}
L.~Hoegner and U.~Stilla, ``Automatic {3D} reconstruction and texture
  extraction for {3D} building models from thermal infrared image sequences,''
  \emph{Quant. InfraRed Thermogr}, 2016.

\bibitem{guidi_multi-resolution_2009}
\BIBentryALTinterwordspacing
G.~Guidi, M.~Russo, S.~Ercoli, F.~Remondino, A.~Rizzi, and F.~Menna, ``A
  {Multi}-{Resolution} {Methodology} for the {3D} {Modeling} of {Large} and
  {Complex} {Archeological} {Areas},'' \emph{International Journal of
  Architectural Computing}, vol.~7, no.~1, pp. 39--55, Jan. 2009, publisher:
  SAGE Publications. [Online]. Available:
  \url{https://doi.org/10.1260/147807709788549439}
\BIBentrySTDinterwordspacing

\bibitem{qin_rpc_2016}
R.~Qin, ``Rpc stereo processor ({RSP})–a software package for digital surface
  model and orthophoto generation from satellite stereo imagery,''
  \emph{International Archives of the Photogrammetry, Remote Sensing \& Spatial
  Information Sciences}, vol.~3, p.~77, 2016, publisher: Copernicus GmbH.

\bibitem{qin_automated_2017}
------, ``Automated {3D} recovery from very high resolution multi-view images
  {Overview} of {3D} recovery from multi-view satellite images,'' in
  \emph{{ASPRS} {Conference} ({IGTF}) 2017}, 2017, pp. 12--16.

\bibitem{waechter_let_2014}
M.~Waechter, N.~Moehrle, and M.~Goesele, ``Let there be color! {Large}-scale
  texturing of {3D} reconstructions,'' in \emph{European conference on computer
  vision}.\hskip 1em plus 0.5em minus 0.4em\relax Springer, 2014, pp. 836--850.

\bibitem{lempitsky_seamless_2007}
V.~Lempitsky and D.~Ivanov, ``Seamless mosaicing of image-based texture maps,''
  in \emph{2007 {IEEE} conference on computer vision and pattern
  recognition}.\hskip 1em plus 0.5em minus 0.4em\relax IEEE, 2007, pp. 1--6.

\bibitem{bi_patch-based_2017}
\BIBentryALTinterwordspacing
S.~Bi, N.~K. Kalantari, and R.~Ramamoorthi,
  ``\BIBforeignlanguage{en}{Patch-based optimization for image-based texture
  mapping},'' \emph{\BIBforeignlanguage{en}{ACM Trans. Graph.}}, vol.~36,
  no.~4, pp. 1--11, Jul. 2017. [Online]. Available:
  \url{https://dl.acm.org/doi/10.1145/3072959.3073610}
\BIBentrySTDinterwordspacing

\bibitem{callieri_masked_2008}
\BIBentryALTinterwordspacing
M.~Callieri, P.~Cignoni, M.~Corsini, and R.~Scopigno,
  ``\BIBforeignlanguage{en}{Masked photo blending: {Mapping} dense photographic
  data set on high-resolution sampled {3D} models},''
  \emph{\BIBforeignlanguage{en}{Computers \& Graphics}}, vol.~32, no.~4, pp.
  464--473, Aug. 2008. [Online]. Available:
  \url{https://www.sciencedirect.com/science/article/pii/S009784930800054X}
\BIBentrySTDinterwordspacing

\bibitem{yanover_finding_2003}
C.~Yanover and Y.~Weiss, ``Finding the {M} most probable configurations using
  loopy belief propagation,'' \emph{Advances in neural information processing
  systems}, vol.~16, pp. 289--296, 2003, publisher: Citeseer.

\bibitem{kolmogorov_what_2004}
V.~Kolmogorov and R.~Zabin, ``What energy functions can be minimized via graph
  cuts?'' \emph{IEEE Transactions on Pattern Analysis \& Machine Intelligence},
  vol.~26, no.~2, pp. 147--159, 2004, publisher: IEEE.

\bibitem{paragios_graph_2006}
\BIBentryALTinterwordspacing
Y.~Boykov and O.~Veksler, ``\BIBforeignlanguage{en}{Graph {Cuts} in {Vision}
  and {Graphics}: {Theories} and {Applications}},'' in
  \emph{\BIBforeignlanguage{en}{Handbook of {Mathematical} {Models} in
  {Computer} {Vision}}}, N.~Paragios, Y.~Chen, and O.~Faugeras, Eds.\hskip 1em
  plus 0.5em minus 0.4em\relax New York: Springer-Verlag, 2006, pp. 79--96.
  [Online]. Available: \url{http://link.springer.com/10.1007/0-387-28831-7_5}
\BIBentrySTDinterwordspacing

\bibitem{li_markov_1994}
S.~Z. Li, ``\BIBforeignlanguage{en}{Markov random field models in computer
  vision},'' in \emph{\BIBforeignlanguage{en}{Computer {Vision} — {ECCV}
  '94}}, ser. Lecture {Notes} in {Computer} {Science}, J.-O. Eklundh, Ed.\hskip
  1em plus 0.5em minus 0.4em\relax Berlin, Heidelberg: Springer, 1994, pp.
  361--370.

\bibitem{xu_an_2011}
M.~Xu, H.~Chen, and P.~K. Varshney, ``An image fusion approach based on markov
  random fields,'' \emph{IEEE Transactions on Geoscience and Remote Sensing},
  vol.~49, no.~12, pp. 5116--5127, 2011.

\bibitem{grammatikopoulos_automatic_2007}
L.~Grammatikopoulos, I.~Kalisperakis, G.~Karras, and E.~Petsa, ``Automatic
  multi-view texture mapping of {3D} surface projections,'' in
  \emph{Proceedings of the 2nd {ISPRS} {International} {Workshop}
  {3D}-{ARCH}}.\hskip 1em plus 0.5em minus 0.4em\relax Citeseer, 2007, pp.
  1--6.

\bibitem{allene_seamless_2008}
C.~Allene, J.-P. Pons, and R.~Keriven, ``Seamless image-based texture atlases
  using multi-band blending,'' in \emph{2008 19th {International} {Conference}
  on {Pattern} {Recognition}}, Dec. 2008, pp. 1--4, iSSN: 1051-4651.

\bibitem{baumberg_blending_2002}
A.~Baumberg, ``Blending {Images} for {Texturing} {3D} {Models}.'' in
  \emph{Bmvc}, vol.~3.\hskip 1em plus 0.5em minus 0.4em\relax Citeseer, 2002,
  p.~5.

\bibitem{li_fast_2006}
Y.~Li and L.~Ma, ``A fast and robust image stitching algorithm,'' in \emph{2006
  6th world congress on intelligent control and automation}, vol.~2.\hskip 1em
  plus 0.5em minus 0.4em\relax IEEE, 2006, pp. 9604--9608.

\bibitem{szeliski_image_2006}
R.~Szeliski, ``Image alignment and stitching,'' in \emph{Handbook of
  mathematical models in computer vision}.\hskip 1em plus 0.5em minus
  0.4em\relax Springer, 2006, pp. 273--292.

\bibitem{gal_seamless_2010}
\BIBentryALTinterwordspacing
R.~Gal, Y.~Wexler, E.~Ofek, H.~Hoppe, and D.~Cohen-Or,
  ``\BIBforeignlanguage{en}{Seamless {Montage} for {Texturing} {Models}},''
  \emph{\BIBforeignlanguage{en}{Computer Graphics Forum}}, vol.~29, no.~2, pp.
  479--486, 2010, \_eprint:
  https://onlinelibrary.wiley.com/doi/pdf/10.1111/j.1467-8659.2009.01617.x.
  [Online]. Available:
  \url{https://onlinelibrary.wiley.com/doi/abs/10.1111/j.1467-8659.2009.01617.x}
\BIBentrySTDinterwordspacing

\bibitem{fu_texture_2018}
Y.~Fu, Q.~Yan, L.~Yang, J.~Liao, and C.~Xiao, ``Texture mapping for 3d
  reconstruction with rgb-d sensor,'' in \emph{Proceedings of the {IEEE}
  conference on computer vision and pattern recognition}, 2018, pp. 4645--4653.

\bibitem{li_fast_2018}
W.~Li, H.~Gong, and R.~Yang, ``Fast texture mapping adjustment via local/global
  optimization,'' \emph{IEEE transactions on visualization and computer
  graphics}, vol.~25, no.~6, pp. 2296--2303, 2018, publisher: IEEE.

\bibitem{perez_poisson_2003}
P.~Pérez, M.~Gangnet, and A.~Blake, ``Poisson image editing,'' in \emph{{ACM}
  {SIGGRAPH} 2003 {Papers}}, 2003, pp. 313--318.

\bibitem{cernea_openmvs_2015}
\BIBentryALTinterwordspacing
D.~Cernea, ``Openmvs: {Open} multiple view stereovision,'' 2015. [Online].
  Available: \url{https://github.com/cdcseacave/openMVS}
\BIBentrySTDinterwordspacing

\bibitem{kolmogorov_convergent_2005}
V.~Kolmogorov, ``Convergent tree-reweighted message passing for energy
  minimization,'' in \emph{International {Workshop} on {Artificial}
  {Intelligence} and {Statistics}}.\hskip 1em plus 0.5em minus 0.4em\relax
  PMLR, 2005, pp. 182--189.

\bibitem{fabbri_2d_2008}
R.~Fabbri, L.~D.~F. Costa, J.~C. Torelli, and O.~M. Bruno, ``{2D} {Euclidean}
  distance transform algorithms: {A} comparative survey,'' \emph{ACM Computing
  Surveys (CSUR)}, vol.~40, no.~1, pp. 1--44, 2008, publisher: ACM New York,
  NY, USA.

\bibitem{buehler_unstructured_2001}
\BIBentryALTinterwordspacing
C.~Buehler, M.~Bosse, L.~McMillan, S.~Gortler, and M.~Cohen, ``Unstructured
  lumigraph rendering,'' in \emph{Proceedings of the 28th annual conference on
  {Computer} graphics and interactive techniques}, ser. {SIGGRAPH} '01.\hskip
  1em plus 0.5em minus 0.4em\relax New York, NY, USA: Association for Computing
  Machinery, Aug. 2001, pp. 425--432. [Online]. Available:
  \url{https://doi.org/10.1145/383259.383309}
\BIBentrySTDinterwordspacing

\bibitem{zhou_selection_2021}
G.~Zhou, X.~Bao, S.~Ye, H.~Wang, and H.~Yan, ``Selection of optimal building
  facade texture images from uav-based multiple oblique image flows,''
  \emph{IEEE Transactions on Geoscience and Remote Sensing}, vol.~59, no.~2,
  pp. 1534--1552, 2021.

\bibitem{greene_hierarchical_1993}
N.~Greene, M.~Kass, and G.~Miller, ``Hierarchical {Z}-buffer visibility,'' in
  \emph{Proceedings of the 20th annual conference on {Computer} graphics and
  interactive techniques}, 1993, pp. 231--238.

\bibitem{agisoft_llc_metashape_2021}
\BIBentryALTinterwordspacing
A.~LLC., ``Metashape pro,'' Aug. 2021. [Online]. Available:
  \url{http://www.agisoft.com}
\BIBentrySTDinterwordspacing

\bibitem{eagle_eagle-texturemapping_2021}
\BIBentryALTinterwordspacing
Eagle, ``{EAGLE}-{TextureMapping},'' Aug. 2021, original-date:
  2019-09-10T08:52:25Z. [Online]. Available:
  \url{https://github.com/OneEyedEagle/EAGLE-TextureMapping}
\BIBentrySTDinterwordspacing

\bibitem{wang_multiscale_2003}
Z.~Wang, E.~P. Simoncelli, and A.~C. Bovik, ``Multiscale structural similarity
  for image quality assessment,'' in \emph{The {Thrity}-{Seventh} {Asilomar}
  {Conference} on {Signals}, {Systems} \& {Computers}, 2003}, vol.~2.\hskip 1em
  plus 0.5em minus 0.4em\relax Ieee, 2003, pp. 1398--1402.

\bibitem{sensefly_discover_2021}
\BIBentryALTinterwordspacing
senseFly, ``\BIBforeignlanguage{en-US}{Discover a wide range of drone
  datasets},'' 2021. [Online]. Available:
  \url{https://www.sensefly.com/education/datasets/}
\BIBentrySTDinterwordspacing

\bibitem{moulon_openmvg_2016}
P.~Moulon, P.~Monasse, R.~Perrot, and R.~Marlet, ``Openmvg: {Open} multiple
  view geometry,'' in \emph{International {Workshop} on {Reproducible}
  {Research} in {Pattern} {Recognition}}.\hskip 1em plus 0.5em minus
  0.4em\relax Springer, 2016, pp. 60--74.

\bibitem{zhou_color_2014}
Q.-Y. Zhou and V.~Koltun, ``Color {Map} {Optimization} for {3D}
  {Reconstruction} with {Consumer} {Depth} {Cameras},'' \emph{ACM Transactions
  on Graphics}, vol.~33, Aug. 2014.

\end{thebibliography}
\end{document}